\documentclass[]{interact}
\usepackage{amssymb}
\usepackage{amsmath,amsfonts}
\usepackage{relsize}
\usepackage{mathtools}
\usepackage{algorithm}
\usepackage{algpseudocode}
\usepackage{array}
\usepackage[caption=false]{subfig}
\usepackage{textcomp}
\usepackage{stfloats}
\usepackage{url}
\usepackage{verbatim}
\usepackage{relsize}
\usepackage{graphicx}
\usepackage{bm}
\usepackage{makecell}
\usepackage{cite}
\usepackage{sidecap}
\usepackage[usestackEOL]{stackengine}
\algrenewcommand\algorithmicrequire{\textbf{Input:}}
\algrenewcommand\algorithmicensure{\textbf{Output:}}
\def\BibTeX{{\rm B\kern-.05em{\sc i\kern-.025em b}\kern-.08em
    T\kern-.1667em\lower.7ex\hbox{E}\kern-.125emX}}
\usepackage{balance}
\usepackage{color}
\usepackage[dvipsnames]{xcolor}
\usepackage[normalem]{ulem}
\newcommand{\Add}[1]{\textcolor{blue}{#1}}
\DeclareRobustCommand{\Erase}{\bgroup\markoverwith{\Add{\rule[.5ex]{2pt}{0.4pt}}}\ULon}

\definecolor{DarkBlue}{RGB}{0, 50, 120}
\definecolor{LightBlue}{RGB}{60, 130, 195}
\definecolor{ClearBlue}{RGB}{115, 122, 160}

\begin{document}

\articletype{FULL PAPER}

\title{Learning Stable In-Grasp Manipulation in a Non-Dropping Action Space}

\author{
\name{Ha Thang Long Doan\textsuperscript{a}\thanks{CONTACT Kenji Tahara. Email: tahara@ieee.org}, Hikaru Arita \textsuperscript{b}, Kazuto Nakashima \textsuperscript{c},  and Kenji Tahara \textsuperscript{b}}
\affil{\textsuperscript{a}Department of Mechanical Engineering, Graduate School of Engineering, 744 Motooka Nishi-ku, Kyushu University, Fukuoka, Japan; \textsuperscript{b}Department of Mechanical Engineering, Faculty of Engineering, 744 Motooka Nishi-ku, Kyushu University, Fukuoka, Japan; \textsuperscript{c}Department of Interdisciplinary Informatics, Faculty of Information Science and Electrical Engineering, 744 Motooka Nishi-ku, Kyushu University, Fukuoka, Japan}
}

\maketitle

\begin{abstract}
Traditionally, dexterous manipulation controllers are designed using analytic models constrained by strong assumptions about the hand and the objects being manipulated.
Reinforcement learning (RL) has become another common approach in which skills are explored openly in an end-to-end manner but is inefficient because of unnoticeable instability and conflicts in learning objectives.
This paper attempts to efficiently explore stable and accurate manipulation skills by decomposing dexterous skills into multiple simpler/analyzable components.
Each skill component is subsequently learned with constraints and guidance from classical physics and control theory.
Our work shows that for stable grasp, in-grasp reposition/reorientation with different objects, sensor/motor noise, latency, and frictional conditions, skill learning becomes efficient and stable with prior knowledge from theory.
\end{abstract}


\section{Introduction}\label{Intro}

In the field of robotic dexterous manipulation, the goal of in-hand manipulation tasks is to develop a control method that stably and accurately regulates the hand--object system to the desired states. Traditionally, classical mechanics and control theory have been applied to solve this problem.
This approach requires accurate modeling of the robotic fingers and palm, the object to manipulate, and how they interact with each other at contact points \cite{review_dexterous_manipulation}.
These analytical models contain strong assumptions regarding physical parameters, such as fixed friction, stiffness, and damping properties.
In addition, modeling hand--object contact is challenging, making accurate regulation of each part of the robotic hand to control the object states difficult.
Therefore, the manipulation accuracy varies when these methods are applied to robotic hands and objects with varying physical properties.

Recent studies have increasingly turned to end-to-end reinforcement learning (RL) to avoid relying on complex analytic models and instead explore hand motions that accurately regulate hand--object states \cite{openAI}.
However, learning from scratch requires a long training time to optimize multiple objectives and suffers from a `long-tail' instability problem \cite{DLR1}: the agent occasionally discovers `unstable' actions, such as unnecessarily releasing some fingers or tossing the object.

Our perspective is that, rather than complicating the analytic models, RL is a more effective method for exploring hand motions for accurate manipulation.
However, when the entire dexterous skill is explored with end-to-end learning, the long-tail problem is difficult to overcome.
In particular, unstable actions accurately manipulate hand--object systems but cause unnoticeable instabilities that result in eventual object dropping \cite{DLR1}.
When such a phenomenon occurs, the learning process becomes inefficient because of the difficult-to-identify cause of failure and the conflicting RL objectives.

While existing research attempts to fix instability through RL algorithm selection, reward shaping, or curriculum design \cite{DLR1, stable_learning}, we argue that stability should not be left to RL optimization alone.
Therefore, we shift the perspective by decoupling desired dexterous skills into simpler, analyzable components.
Each component is then learned with the guidance of classical physics and control theory.
This is known as guiding RL with scientific knowledge for sample efficiency \cite{physicsRL}-the knowledge guides the exploration of the action space for higher performance and better convergence.

In this paper, the in-grasp manipulation components, where the contact between the fingertips and the object is maintained, are targeted.
We study how to guide RL using the theory behind stability in fingers--thumb opposability-based dynamic grasping (FTODG) \cite{hcr_theory}.
Our contributions are as follows:

\begin{enumerate}
\item As stability is fundamental to safety, we draw on safe RL \cite{safeRL} to develop an FTODG-based `stable action space'.

\item We propose theoretically stable in-grasp learning (TSIGL), leveraging control barrier functions (CBFs) to constrain exploration within the stable action space.

\item We demonstrate that TSIGL framework can prevent object dropping and unstable finger motions, making it more sample efficient than baseline RL methods are.
In-grasp reposition/reorientation skills are learned under varying environmental conditions, and they are more accurate than skills controlled only with FTODG are.
\end{enumerate}

\section{Related Work}\label{Related}
\vspace{0mm}
\subsection{End-to-end Learning of Dexterous Manipulation}

Using end-to-end RL to learn dexterous manipulation skills has become popular.
In recent work, OpenAI \textit{et al.} \cite{openAI} use visual data and Funabashi \textit{et al.} \cite{tactile_learning_manipulation} use tactile data to learn dexterous manipulation.
External sensorless in-hand manipulation skills have also been learned \cite{DLR1, DLR2}, allowing object manipulation using only internal joint sensor feedback at deployment.
More complex tasks, such as solving Rubik's Cube, have also been presented \cite{openAI_rubik}.
In these frameworks, the RL agents decide the desired joint states to perform the task and then let the joint-level controllers compute the robot's final control signals.
By using joint-level controllers based on control theory (such as PD position control or impedance control), strict joint safety limits are applied.
Hand--object stability is also an important aspect of safety in manipulation.
However, it is still softly promoted, resulting in the long-tail instability problem.
Tao \textit{et al.} \cite{stable_learning} introduce a learning framework with stability being softly encouraged on the basis of heuristic criteria such as `more contact points' and `keep the object in the palm area'.
However, because of attempts to learn multiple dexterous skills at once, determining more specific, theoretically derived criteria is difficult.
Sievers \textit{et al.} \cite{DLR1} apply a simulation environment with a gravity curriculum and reward shaping to promote force closure, a fundamental property for ensuring object stability in the hand \cite{robot_book}.
However, the agent sometimes sacrifices stability to change the object states, resulting in the hand dropping the object.

By applying the theory behind stability and targeting in-grasp manipulation skills specifically, our guided RL method achieves better sample efficiency than learning from scratch.

\subsection{Robotic Control Theories for Dexterous Manipulation}

The control theories behind how humans manipulate objects have been studied, such as power grasp to wrap an object using the whole hand \cite{power_grasp}, in-hand relocation \cite{position_control}, in-hand reorientation \cite{analytical_reorientation}, or sliding regrasp and relocation \cite{slide_regrasp}.
Control theories of how to manipulate objects in the dark or external sensorless stable grasp \cite{hcr_stable_grasping}, relocation and reorientation \cite{hcr_virtual_frame}, and regrasp \cite{hcr_regrasp} have also been studied.
In general, this field studies predefined sets of rules to compute actions that theoretically can regulate hand--object contact to the desired states.
For this reason, suitable theories can be applied so that the actions are as stable as mathematically proven.
However, the robot, object, and contact models in the proofs are constrained by many assumptions \cite{review_dexterous_manipulation, review_dexterous_manipulation_PROBLEM, review_analytical-based}, such as one specific type of contact, fixed joint friction, stiffness, and damping properties, resulting in inaccurate manipulation.

Our standpoint is that control theories are crucial in stable manipulation, but their model assumptions are the bottleneck, and RL can be utilized to address such challenges.

\subsection{RL Guided by Scientific Knowledge}

Combining learning and scientific knowledge has become a growing field \cite{physicsRL}, with some noticeable methods for robotic manipulation, including physics-informed RL \cite{physicsRL} and residual RL \cite{residual}.
Scientific knowledge can be incorporated softly in the cost function and the environment \cite{RLphysics_reward}, more strictly in the control system model \cite{RLphysics_control}, or with hard constraints at the final filter before being input to the robot \cite{RL_CBFs}.
There are different levels of safety in terms of robotic control, ranging from no to strict requirements.
Control theory leverages dynamic models of the system to obtain a guarantee, while RL usually gives up on a formal guarantee to be adaptable to new contexts \cite{safeRL}.
Therefore, integrating control theories to satisfy safety requirements is a promising approach for safe RL \cite{safeRL}.

In this paper, we study how to incorporate control theories to constrain strict hand--object stability requirements.

\section{Proposal}
\subsection{FTODG}\label{applied_theories}

In this section, we introduce the control theory of fingers--thumb opposability-based dynamic grasping (FTODG) \cite{hcr_theory} and its extended work \cite{hcr_stable_grasping, hcr_virtual_frame, hcr_regrasp, superposition, hcr_object_ori_feedback, hcr_tactile_manipulation, hcr_long}.

\vspace{1mm}
\subsubsection{\textbf{Stable Grasp}}

FTODG first studies control design to stably grasp objects on the basis of the following concepts:

\paragraph{Virtual Frame (VF)}

Let set $I$ consist of at least 3 fingertips $i$ that maintain contact with the object (see Figure.~\ref{VF}(a)).
The hand--object system is considered a VF generated by connecting $\bm{X}_i$, the coordinates of the fingertips $i$ \cite{hcr_virtual_frame}.
The VF position is the weighted centroid of the frame:
\begin{align}
\bm{X}_\textrm{VF} = \dfrac{\sum{ f_{\textrm{d}\_i}\bm{X}_i}}{\sum{f_{\textrm{d}\_i}}}, \ \ \ i \in I \subseteq \{\textrm{th, ff, mf, rf, lf}\},
\end{align}
where $f_{\textrm{d}\_i}$ represents the weights and where th, ff, mf, rf, and lf are the names of the fingers.

\paragraph{Force/Torque Equilibrium Grasp}

By applying the VF concept, the signals $\bm{u}_{\textrm{stable}\_i}$ to stabilize the object can be computed using only internal joint positions:
\begin{equation}
\bm{u}_{\textrm{stable}\_i} = f_{\textrm{d}\_i} \bm{J}_i^\textrm{T}(\bm{X}_\textrm{VF} - \bm{X}_i) - \operatorname*{diag}(\bm{C}_{i}) \dot{\bm{q}}_{i},
\label{u_stable}
\end{equation}
where $\bm{J}_i^\textrm{T}$ is the transposed Jacobian matrix with respect to the fingertip.
The first term on the right-hand side represents the object \textit{force/torque equilibrium}, which is another fundamental property for ensuring object stability.
The magnitudes of $f_{\textrm{d}\_i}$ determine the nominal stable grasp forces, which influence both the maximum weight the hand can stably grasp and the amount of external force it can withstand.
The ratio of $f_{\textrm{d}\_i}$ determines not only the VF position but also where the fingertips exert a controlled force to stabilize the object (see Figure.~\ref{VF}(a)).
The second term on the right-hand side is the multiplication of the joint velocity $\dot{\bm{q}}_{i}$ and damping $\bm{C}_{i}$, resulting in joint damping.

\subsubsection{\textbf{Stable In-Grasp Manipulation}}

Once the object is stabilized, FTODG studies control design to stably manipulate grasped object states while maintaining fingertip contacts:

\paragraph{Principle of Manipulation Signal Superposition}

The \textit{principle of superposition} \cite{superposition} proves that the superposition of the stable grasp and reposition/reorientation torque signals allows the regulation of hand--object states while keeping the object stable in the hand.
Therefore, to in-grasp manipulate the hand--object system, the final signal $\bm{u}_{i}$ is the superposition of the stable grasp term $\bm{u}_{\textrm{stable}\_i}$, the position manipulation $\bm{u}_{\textrm{pos}\_i}$, and the orientation manipulation $\bm{u}_{\textrm{ori}\_i}$ terms:
\begin{equation}
\begin{split}
\bm{u}_{i} = \bm{u}_{\textrm{stable}\_i} + \bm{u}_{\textrm{pos}\_i} + \bm{u}_{\textrm{ori}\_i}.
\end{split}
\end{equation}

\paragraph{Finger Reposition/Reorientation Motions Guided by Feedback Control}

The manipulation control terms for each finger $\bm{u}_{\textrm{pos}\_i}$ and $\bm{u}_{\textrm{ori}\_i}$ are to stably exert the fingertip force/moment for reposition/reorientation, respectively.
Therefore, they are guided by proportional \textit{feedback control} \cite{hcr_theory}. First, the control error is computed (see equations (\ref{e1}) and (\ref{e2})).
Afterward, the desired fingertip forces/moments to stably minimize the error are derived and converted to control signals via force control (see equations (\ref{signal1}) and (\ref{signal2})).
If external sensors, such as cameras, are used for object state feedback, the object states are used directly to compute the control error.
Otherwise, 
VF states are defined, the error is computed using only internal joint sensor data,
and external sensorless in-grasp manipulation is stably realized.
The details are as follows:

\begin{figure}[!t]
\centering
\includegraphics[width=0.85\linewidth]{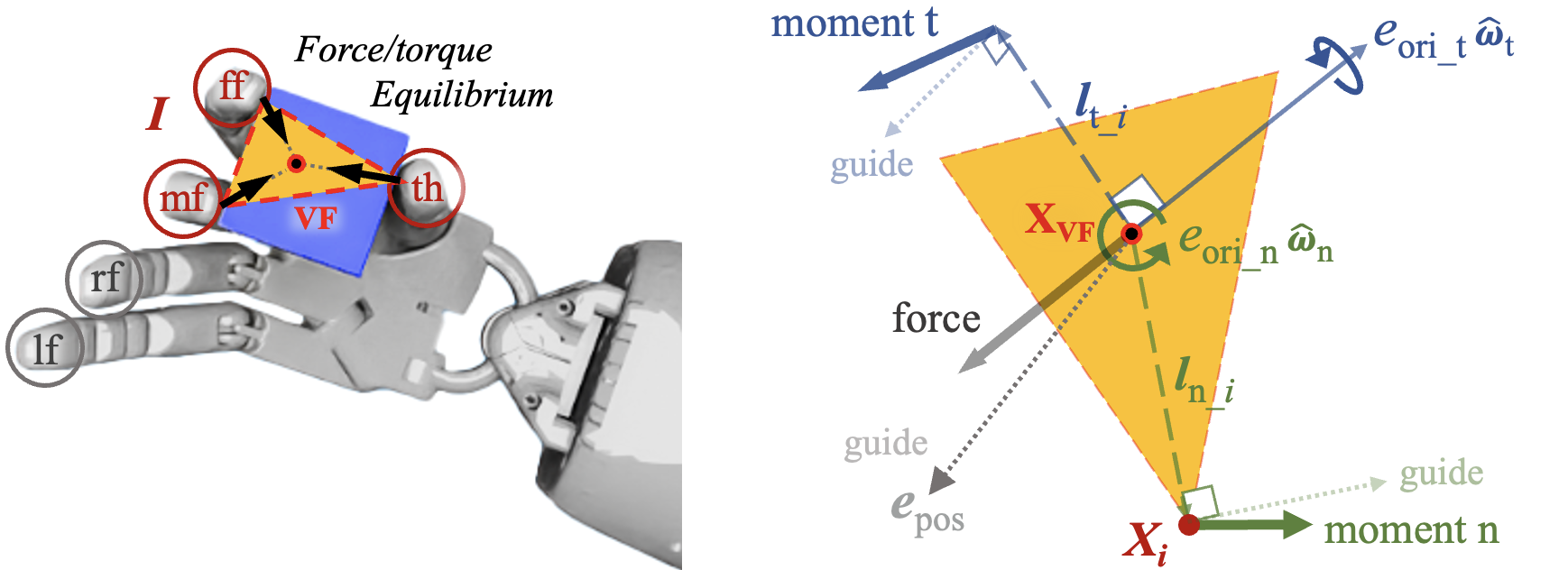}
\\
\vspace{2mm}
(a)  \ \ \ \ \ \ \ \  \ \ \ \ \ \ \ \  \ \ \ \ \ \ \ \  \ \ \ \ \ \ \ \  \ \ \ \ \ \ \ \ \                (b)
\caption{FTODG control design: (a) \textit{Force/torque equilibrium} achieves object stability, while (b) \textit{feedback control} error guides the fingertip force/moment for stable reposition/reorientation.
}
\vspace{-2mm}
\label{VF}
\end{figure}

Reposition: Let $\bm{X}_\textrm{d}$ be the desired position, the current VF position be $\bm{X}_\textrm{VF}$, and the real object position be $\bm{X}_\textrm{object}$.
Then, the \textit{feedback control} signals $\bm{u}_{\textrm{pos}\_i}$ for each finger $i$ to stably minimize the position error $\bm{e}_{\textrm{pos}}$ are straightforward:
\begin{align}
\bm{e}_{\textrm{pos}} = \ \bm{X}_\textrm{d} + \underbrace{(\bm{X}_\textrm{VF} - \bm{X}_\textrm{object})}_{\textrm{available; otherwise, 0}} - \bm{X}_\textrm{VF} \label{e1},\\[1mm]
\bm{u}_{\textrm{pos}\_i} = \bm{J}_i^\textrm{T} \ \bigl[\ \operatorname*{diag}(\bm{P}_{i}) \bm{e}_{\textrm{pos}} \
\bigr],\label{signal1} \ \ \ \ \ \ \ \ 
\end{align}
where $\bm{P}_i$ is the positive gain vector that determines the fingertip force for repositioning, guided by the position control error (shown as gray arrows in Figure.~\ref{VF}(b)).

Reorientation: The VF orientation $\bm{R}_{\textrm{VF}} = [\bm{r}_{\textrm{VF}x}, \bm{r}_{\textrm{VF}y}, \bm{r}_{\textrm{VF}z}]$ is defined using the positions of the internal joints.
In this paper, the method in \cite{hcr_object_ori_feedback} is based on defining the orientation.
Let the desired orientation be $\bm{R}_{\textrm{d}}$ and the current object orientation be $\bm{R}_{\textrm{object}}$; then, the orientation error $e_\textrm{ori}\bm{\hat\omega}$ is calculated as follows:
\begin{align}
& \ \ [\bm{r}_{\textrm{d}x}, \bm{r}_{\textrm{d}y}, \bm{r}_{\textrm{d}z}] = \bm{R}_{\textrm{d}}\underbrace{\bm{R}_{\textrm{object}}^{-1}\bm{R}_{\textrm{VF}}}_{\mathclap{\textrm{available; otherwise, Identity matrix}}},\\
\begin{split}
e_\textrm{ori}\bm{\hat\omega} &= ([\bm{r}_{\textrm{VF}x}\times]\bm{r}_{\textrm{d}x} + [\bm{r}_{\textrm{VF}y}\times] \bm{r}_{\textrm{d}y} + [\bm{r}_{\textrm{VF}z}\times] \bm{r}_{\textrm{d}z})\\
&= \ e_\textrm{ori\_n}\ \bm{\hat\omega}_\textrm{n} + e_\textrm{ori\_t}\ \bm{\hat\omega}_\textrm{t}
\end{split},\label{e2}
\end{align}
where $[.\times]$ is used to obtain a skew-symmetric matrix; hence, the sum of all cross products between the current VF and the desired VF axes is used to compute the error $e_\textrm{ori}\bm{\hat\omega}$.
The error is then decomposed using $\bm{\hat\omega}_\textrm{n}$-a unit vector that is normal to the VF surface, pointing away from the palm, and $\bm{\hat\omega}_\textrm{t}$-the remaining vector that is tangent to it (see Figure.~\ref{VF}(b)).
The detailed decomposition process for equation (\ref{e2}) is explained in \cite{hcr_regrasp}.
Then, the moment arms $\bm{l}_{\textrm{n}\_i}$ and $\bm{l}_{\textrm{t}\_i}$ for each fingertip to generate desired reorientation are derived as follows:
\begin{align}
&\bm{l}_{\textrm{n}\_i} = (\bm{X}_\textrm{VF} - \bm{X}_i) / ||\bm{X}_\textrm{VF} - \bm{X}_i||, \\
&\bm{l}_{\textrm{t}\_i} = [\bm{\hat\omega}_\textrm{n} \times] \bm{\hat\omega}_\textrm{t}.
\end{align}

Finally, the \textit{feedback control} signals $\bm{u}_{\textrm{ori}\_i}$ for each finger $i$ to stably minimize the orientation error are computed:
\begin{align}
\begin{split}
\bm{u}_{\textrm{ori}\_i} &\ = \ \bm{J}_i^\textrm{T}\ \bigl[ \ \ \ \ \ \ \operatorname*{diag}(\bm{O}_{\textrm{n}\_i}) [\bm{l}_{\textrm{n}\_i} \times] e_\textrm{ori\_n} \bm{\hat\omega}_\textrm{n} \\
 & \ \ \ \ \ \ \ \ \ \ \ \ +  \ \ \operatorname*{diag}(\bm{O}_{\textrm{t}\_i}) [\bm{l}_{\textrm{t}\_i} \times] e_\textrm{ori\_t}\ \bm{\hat\omega}_\textrm{t} \ \ \ \ \  \bigr],
\end{split}\label{signal2}
\end{align}
where $\bm{O}_{\textrm{n}\_i}$ and $\bm{O}_{\textrm{t}\_i}$ are the positive gain vectors that determine the moments for reorientation, guided by the orientation control error (shown as green and blue arrows in Figure.~\ref{VF}(b)).

\subsection{Stable Action Space}\label{properties}

In FTODG control design, the proofs of grasp stability and controlled system stability are grounded in Newtonian/Lagrangian mechanics and passivity-based nonlinear control theory \cite{hcr_theory, hcr_stable_grasping, hcr_virtual_frame, superposition}.
Each mathematical proof is constrained to a specific controller with fixed parameters and a specific hand--object system with fixed physical properties, simplified dynamics, and rolling contact models.
Then, real-world experiments \cite{hcr_regrasp, hcr_object_ori_feedback, hcr_tactile_manipulation, hcr_long} confirm the followings:

\subsubsection{\textbf{Theoretical Stability}}\label{theoretical_stability}
The FTODG control design $u_i$ is confirmed to be stable from the following aspects:

\begin{itemize}
\item Dynamic stability: Under dynamic conditions involving external forces, the control signals $u_i$ stably regulate the hand--object system while maintaining a stable grasp.
This is applicable to various robotic hands and objects, not only the exact scenarios in the mathematical proofs.

\item External sensorless: In dexterous manipulation, external sensor data are prone to noise and are occasionally unavailable because of camera lag \cite{hcr_object_ori_feedback} or occlusions of motion capture markers \cite{openAI}.
Only the internal sensors of the robot are necessary to compute $u_i$, and the robots are operated stably in unstructured dynamic environments.

\item Customizability: The $u_i$ components, the selection of fingers to use, and external sensor feedback usage can be analytically customized on the basis of the skill (see Table~\ref{table}) and the robot configuration.

\end{itemize}

\begin{figure}[!t]
\centering
\includegraphics[width=0.75\linewidth]{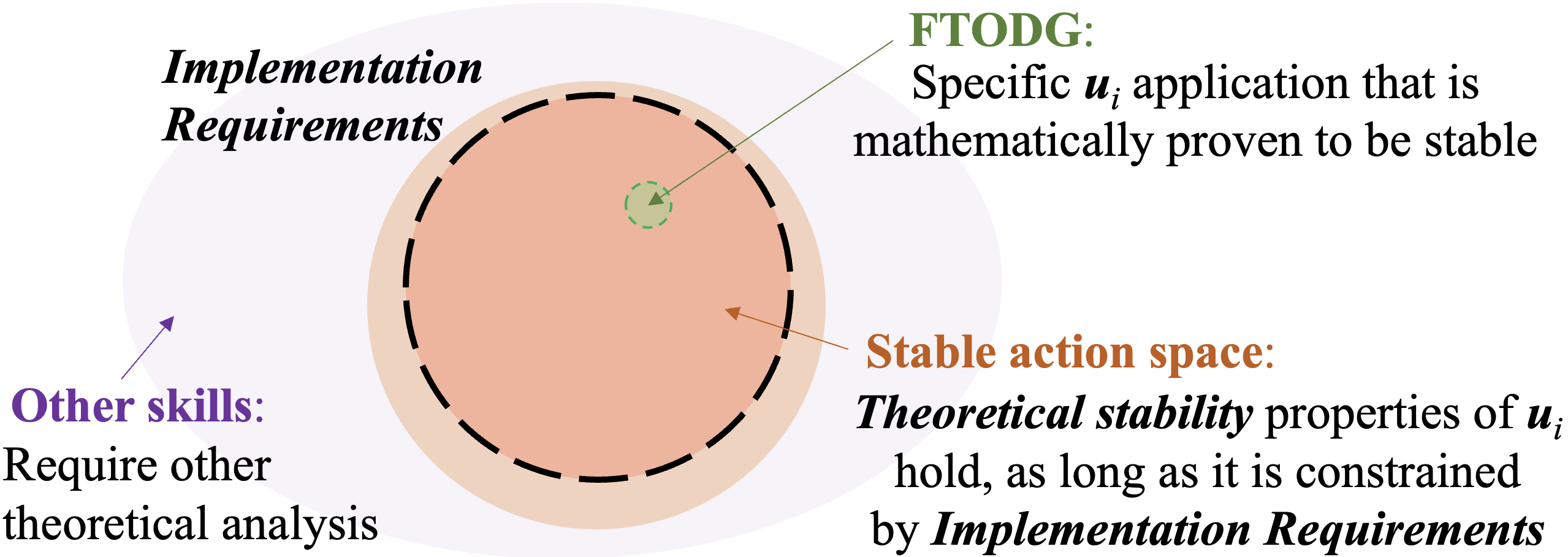}
\vspace{-3mm}
\caption{Stable action space for in-grasp manipulation learning.}
\label{space}
\vspace{-2mm}
\end{figure}

Overall, the $u_i$ design strength of FTODG is the theory behind stability.
$u_i$ is customizable, and the above stability properties hold in the scenarios that are close but beyond what have been mathematically proven.
We refer to this as the FTODG-based `stable action space' (see Figure.~\ref{space}).
Exploration in this space is considered promising for safe RL \cite{safeRL}.

\subsubsection{\textbf{Accuracy Limitations}}\label{limitations}

Owing to the variation in hand--object properties, inaccuracy in hand--object state regulation is a weakness of the FTODG's control design $u_i$.
Therefore, exploring fingertip force/moment for accurate manipulation within FTODG-based stable action space is necessary.

\vspace{1mm}
\subsubsection{\textbf{Implementation Requirements}}\label{requirements}
To construct a stable action space so that the stability properties of FTODG hold, the following requirements are important (see Figure.~\ref{space}):
\begin{itemize}
\item Robotic hands and objects: The theory is applicable to torque-controlled hands that operate within their manipulation workspace to execute the control signals $u_i$.
The objects to manipulate should be rigid, whose shapes and materials result in minimal slippage with the fingertips.

\item Fingertip force/moment: The stable grasp signals during manipulation are required not to be overshadowed by the reposition/reorientation signals in the \textit{principle of superposition};
That is, first, a significant change in \textit{feedback control} gains during manipulation should be constrained.
Second, the \textit{force/torque equilibrium} should be constrained to be above $F_{\textrm{min}\_i}$ to hold the object tightly and below $F_{\textrm{max}\_i}$ to avoid excessive force on the object:
\begin{align}
h_{\textrm{min}\_i} = \textrm{proj}_{(\bm{X}_\textrm{VF} - \bm{X}_i)}\bm{F}_i - F_{\textrm{min}\_i} \geq 0,\label{barrier1}\\
h_{\textrm{max}\_i} = F_{\textrm{max}\_i} - \textrm{proj}_{(\bm{X}_\textrm{VF} - \bm{X}_i)}\bm{F}_i \geq 0,
\label{barrier2}
\end{align}
where $\bm{F}_i$ is the fingertip force exerted on the object and where $\textrm{proj}_{\bm{B}}\bm{A}$ is the scalar projection of vector $\bm{A}$ on $\bm{B}$.
\end{itemize}

\subsection{TSIGL}\label{method}

\begin{figure}[!t]
\centering
\includegraphics[width=0.85\linewidth]{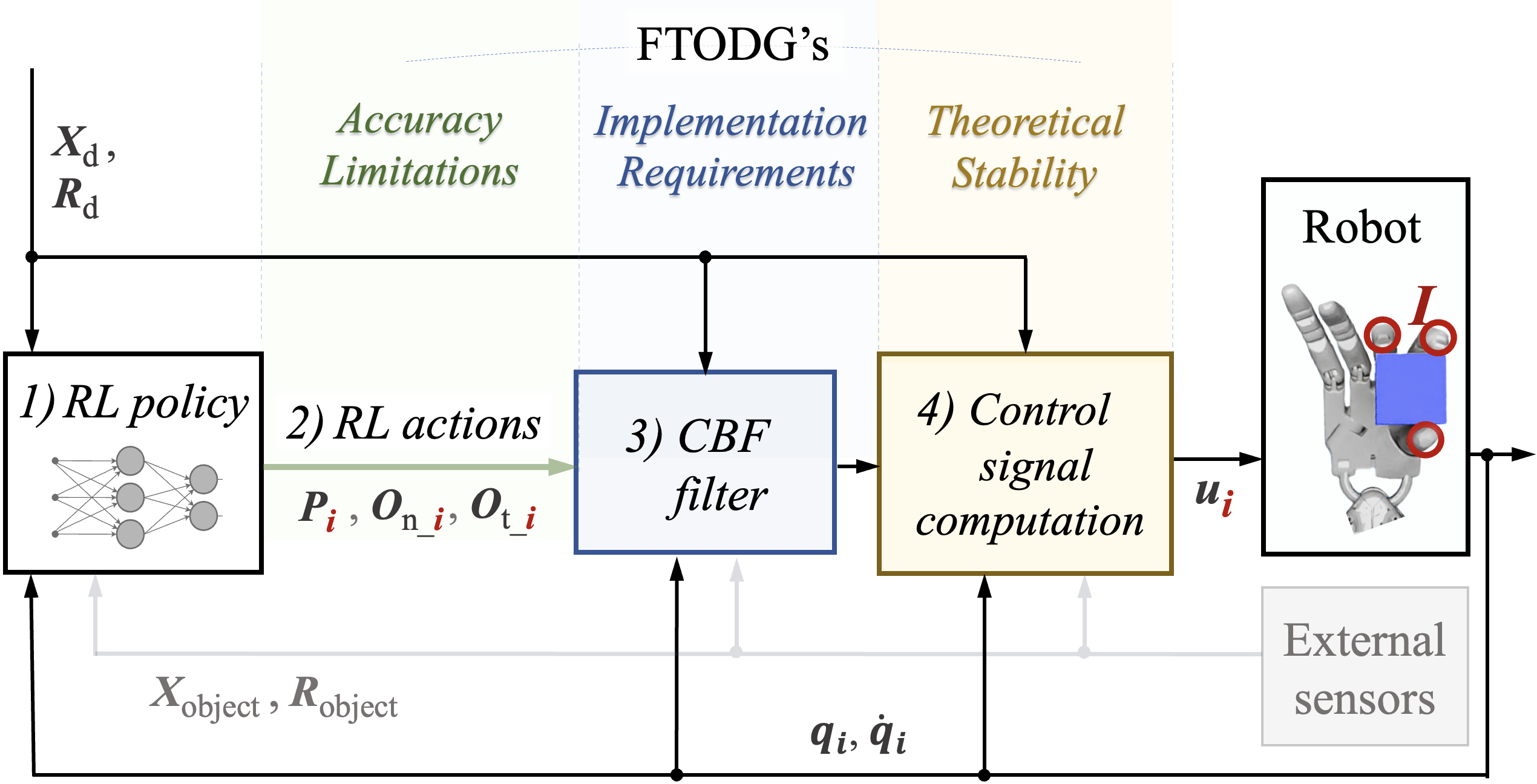}
\vspace{-2mm}
\caption{TSIGL framework: An RL policy learns stable manipulation skills by utilizing the strength (stability) of FTODG and learns to solve its limitations (accuracy).
The framework leverages CBFs to filter actions to satisfy FTODG's implementation requirements, constraining the stable action space.}
\label{VF2}
\end{figure}

\begin{table}[!t]
\def\tablename{Table}
\centering
\caption{Customized integration of FTODG, RL actions, and visual sensor information on the basis of the skill to learn}
\vspace{3mm}
\includegraphics[width=0.8\linewidth]{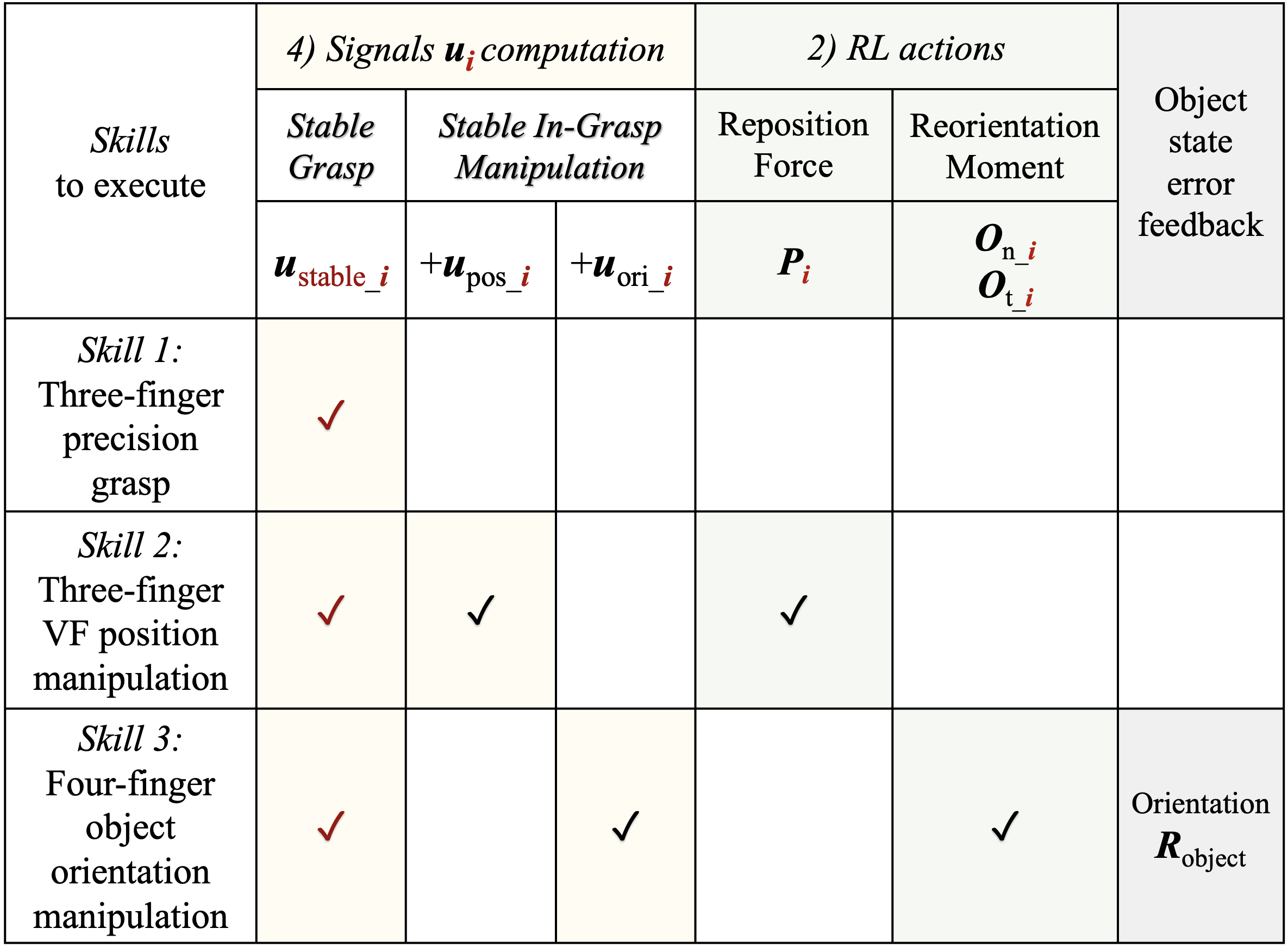}
\vspace{-2mm}
\label{table}
\end{table}

On the basis of the analysis in section \ref{properties}, we propose TSIGL framework (see Figure.~\ref{VF2} and Table~\ref{table}) to learn stable in-grasp manipulation skills:

\subsubsection{RL Policy}
First, the observation states $\bm{s}$ are collected for the RL policy $\pi^{\textrm{RL}}(\bm{a}|\bm{s})$ to map them to actions $\bm{a}$.
\begin{equation}
\bm{a}  \ \ \  \sim \ \ \  \pi^{\textrm{RL}}(\bm{a}|\bm{s}).
\label{scale}
\end{equation}

\subsubsection{RL Actions}
As explained in section \ref{limitations}, inaccuracy in repositioning/reorientation is the limitation of FTODG.
Therefore, the RL actions are set as elements of $\bm{P}_i$, $\bm{O}_{\textrm{n}\_i}$, and $\bm{O}_{\textrm{t}\_i}$ for the fingertip force/moment to be explored for accurate manipulation.
Actions $\bm{a}$ are linearly scaled to the limit range of each element $[K_{\textrm{min}}^k, K_{\textrm{max}}^k]$ to obtain $\bm{K}_{\textrm{RL}}$.

\subsubsection{CBF Filter}
For constrained exploration within the stable action space, the actions need to meet the FTODG implementation requirements in section \ref{requirements} (see Figure.~\ref{space}).
Some requirements are simple: the selection of the action limit $K_{\textrm{max}}^k$ constrains the workspace limit, and the weighted moving average filter constrains the frequent change in the actions $\bm{a}$.
However, the requirements to prevent fingers $i$ from breaking contact and from exerting excessive force (see inequalities $h_{\textrm{min}\_i}$ (\ref{barrier1}) and $h_{\textrm{max}\_i}$ (\ref{barrier2})) are inequalities of linear combinations of action vector components, which are difficult to constrain using common RL techniques.
The inequalities have the form of \textit{control barrier functions} (\textit{CBFs}), an advanced technique that has been proven to be effective in safe RL \cite{safeRL}.
Therefore, we develop a filter based on CBFs to constrain $\bm{K}_{\textrm{RL}}$ to $\bm{K}$ that satisfies the FTODG implementation requirements:

First, we apply a simple static model of the finger as follows:
Assuming that the fingertip exerts force on the object while not causing significant internal motions between the links (static/quasistatic condition) \cite{robot_book}, then the force that will be exerted on the object $\bm{F}_{i}$ at $t+1$ can be roughly estimated from the controller torque signals $\bm{u}_{i}$ at $t$:
\begin{equation}
\bm{F}_{i, t+1} = \bm{J}_{i, t}^\textrm{-T} \bm{u}_{i, t}(\bm{K}_{t}, \bm{s}_{t}).
\label{invert}
\end{equation}
This derivation is based on the \textit{principle of virtual work} in mechanics.
Accordingly, the actions in vector form, $\bm{K}_{t}$, input to the system at hand--object state $\bm{s}_{t}$ outputs fingertip force vector $\bm{F}_{i, t+1}$.
A simple static model computed with noisy hand--object state $\bm{s}_{t}$ observations (explained in section~\ref{env}) is applied in this paper to demonstrate the applicability of the method when accurate information is not available.
Under such limitations, the fingertip force limits $F_{\textrm{min}\_i}$ and $F_{\textrm{max}\_i}$ in the constraints are simply set to be stricter.

The CBF filter is formulated as follows:
\begin{align}
\bm{K} = \bm{K}_{\textrm{RL}}(\bm{s}) + \bm{K}_{\textrm{CBF}}(\bm{s}, \bm{K}_{\textrm{RL}}),
\end{align}
where $\bm{K}_{\textrm{CBF}}$ is obtained by solving a quadratic programming problem so that it can `push' the unconstrained part of $\bm{K}_{\textrm{RL}}$ back to the safe set with minimum adjustment.
This process is conducted immediately after $\bm{K}_{\textrm{RL}}$ is computed and immediately before it is input into the FTODG control design, satisfying its implementation requirements both during training and at deployment:
\begin{equation}
\begin{split}
&(\bm{K}_{\textrm{CBF}}, \epsilon) =  \operatorname*{argmin}_{\bm{K}_{\textrm{CBF}}, \epsilon} (\bm{K}_{\textrm{CBF}}^\textrm{T}\bm{H}\bm{K}_{\textrm{CBF}} + C_{\epsilon} \epsilon), \ \ \
\\ &\textrm{s.t.} \ h_{\textrm{min}\_i}(\bm{F}_{i, t+1}) \geq - \epsilon, \ h_{\textrm{max}\_i}(\bm{F}_{i, t+1}) \geq - \epsilon, 
\\& \ \ \ \ \ \ \ \ \ \ \ \ \ K^k \in [K_{\textrm{min}}^k, K_{\textrm{max}}^k],
\end{split}
\label{quadratic_programming}
\end{equation}
where the matrix $\bm{H}$ represents the weights for the `push' between the elements inside $\bm{K}_{\textrm{RL}}$.
$\epsilon$ is a slack variable in the safety condition, and the violation is heavily penalized by a large constant $C_{\epsilon}$.
The quadratic programming formulation is always solvable by permitting a highly conservative action; specifically, $K_{\textrm{min}}^k$ is set to zero, permitting the CBFs to stop manipulation force/moment completely if necessary.

\subsubsection{Control Signal Computation}
Finally, the actions that satisfy the implementation requirements are used to compute control signals $\bm{u}_i$ to the robot.
With $\bm{u}_i$ design, exploration is within the stable action space, as explained in section \ref{theoretical_stability}.

\section{Evaluation}\label{eval}
To evaluate our TSIGL framework in realizing stable and accurate manipulation, the following skills are executed/learned:

\begin{itemize}
\item \textit{Skill 1}: Three-finger precision grasp

\item \textit{Skill 2}: Three-finger VF (external sensorless) position manipulation

\item \textit{Skill 3}: Four-finger object orientation manipulation
\end{itemize}

The integration of the theory, actions to explore, and external sensor usage is summarized in Table~\ref{table}.
A simulation based on the NVIDIA Isaac Lab framework \cite{orbit, IsaacLab} is used to learn manipulation skills in parallel environments.

\subsection{ RL Simulation Setup}\label{sim_setup}

\subsubsection{Environment}\label{env}
96 parallelized instances of the fully actuated torque-controlled shadow dexterous hand \cite{shadow_hand} are used to evaluate our TSIGL framework.
The objects used for training are selected to have basic shapes, including cubes, cuboids, cylinders, and cones of varying sizes.
A tuna can, a spam can, a triangular prism, and a mustard bottle are used to evaluate the learning results.
The robotic hand model, as well as the object properties and goal states, are determined on the basis of the FTODG implementation requirements in section \ref{requirements}.
The gravity vector is pointed away from the palm of the robotic hand to learn a stable grasp using the fingertips.
Static and kinetic friction between the fingertips and the object are randomized with Gaussian noise.
The joints have randomized friction, stiffness, and damping properties that resist the internal motions between the links for the fingertips to exert force on the object \cite{robot_book}.
Random noise and latency for RL observations and actions are added to model backlash, signal transmission errors, and environmental uncertainty.

\vspace{0mm}
\subsubsection{Timing}
Each episode starts with the appearance of the Shadow Hand's fingertips in contact with various random objects with altered positions and postures, after which the agent learns the manipulation task and ends when the maximum consecutive manipulation goals of 50 are achieved, the object is removed from each finger's reach, or at time limit $T$ (seconds) \cite{openAI}.
The manipulation goal is randomized every time the Euclidean distance or angle between the current and desired states is maintained within $\delta$ (meters/radians, depending on the skill) for more than 1.5 seconds.

\vspace{0mm}
\subsubsection{Advantage Actor--Critic RL}\label{PPO}
The general advantage estimator \cite{GAE} is used to approximate the advantage. To apply this method, supervised learning is used to fit the value function with the discounted future rewards collected from the simulation.
Proximal policy optimization (PPO) \cite{PPO} is selected as the optimization algorithm for the policy network, which is generally effective for on-policy dexterous manipulation RL to stably update the policy on the basis of the advantage \cite{openAI}.
Symmetric observations are used for value and policy networks, which include the current joint positions and velocities that are scaled to their limit ranges, fingertip positions and velocities, control error and its rate of change, goal states, and the currently executed actions.
If external sensors are used to capture object states, they are also included.
Specific parameters for the PPO algorithm are available on the project website (\url{https://hathanglongdoan.github.io/Learning_stable_in-grasp_manipulation/}).

\begin{figure}[!b]
\vspace{-3mm}
\centering
\begin{minipage}[c]{0.02\textwidth}
(a)
\end{minipage}
\begin{minipage}[c]{0.6\textwidth}
\centering
\hspace{4mm}
\includegraphics[width=0.85\linewidth]{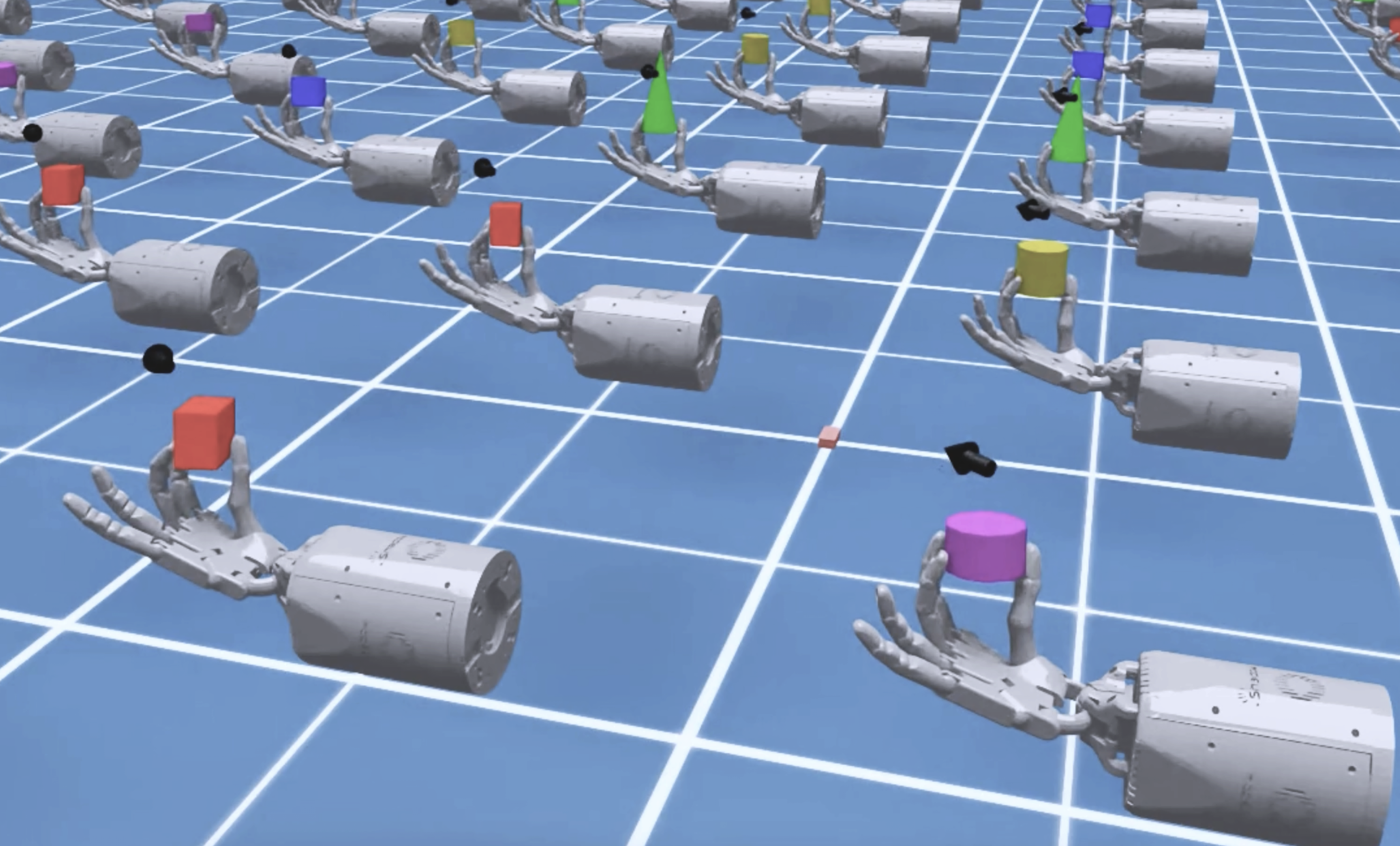}
\end{minipage}
\begin{minipage}[c]{0.30\textwidth}
\renewcommand{\arraystretch}{1.3}
\begin{tabular}{|c | c|}
     \hline
        \makecell{Goal} & \makecell{stable grasp}\\\hline
        \makecell{$I$} & \makecell{th, ff, mf}  \\\hline
         \makecell{Gravity} & \makecell{upward}  \\\hline
\end{tabular}
\end{minipage}
\\
\centering
\vspace{2mm}
\begin{minipage}[c]{0.02\textwidth}
(b)
\end{minipage}
\begin{minipage}[c]{0.7\textwidth}
\centering
\hspace{0mm}
\includegraphics[width=0.95\linewidth]{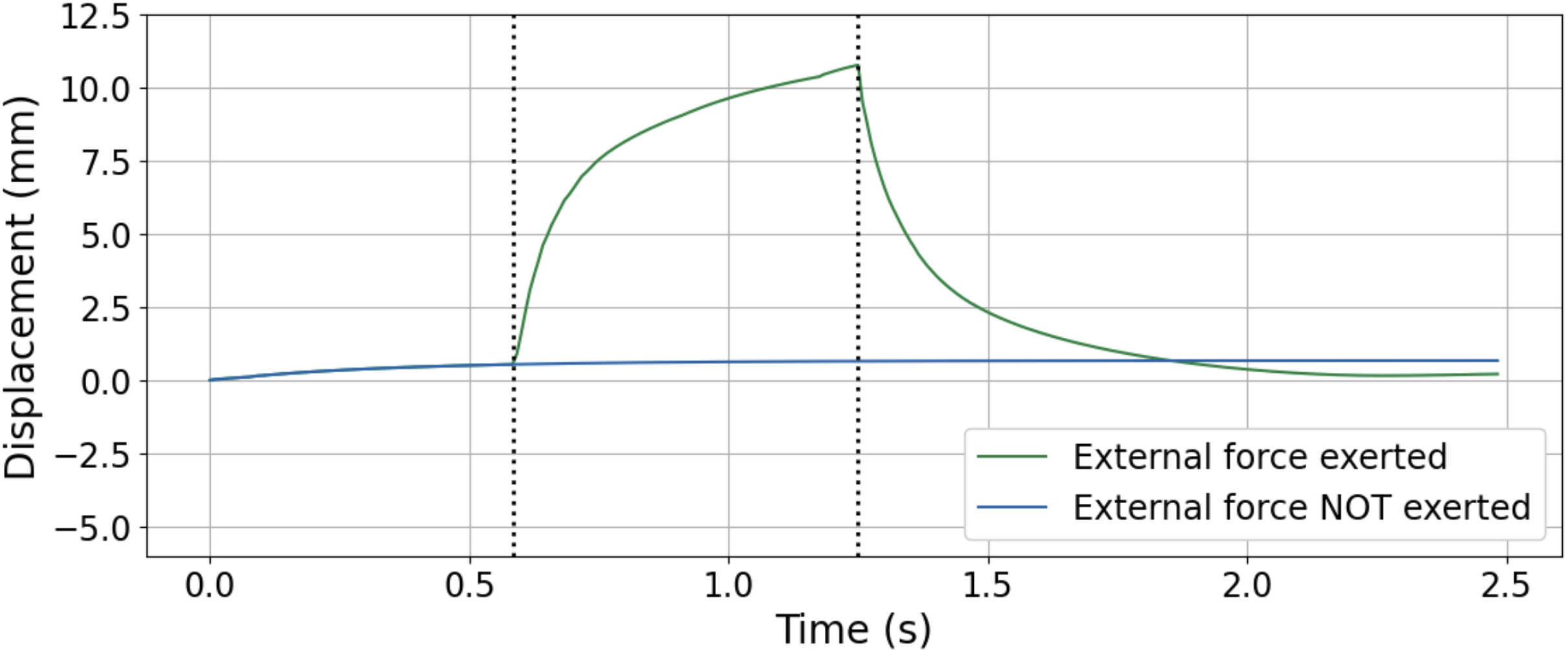}
\end{minipage}
\caption{\textit{Skill 1}: (a) External forces from 0.1 N to 8.6 N applied to objects under \textit{force/torque equilibrium} and (b) displacement of a cuboid when 4.6 N is applied from 0.58 s to 1.25 s.}
\label{precision_grasp}
\end{figure}

\vspace{0mm}
\subsubsection{Reward Function}
The reward function is as follows:
\begin{equation}
r_t = \mathlarger{\sum} \big[\lambda_k|e_k| + r_{\textrm{maintain}\_k} \big]  + r_{\textrm{success}} + r_{\textrm{drop}},
\end{equation}
where $\lambda_k$ is the penalty per meter for each component $k = x, y, z$ of the position error vector $\bm{e}_{\textrm{pos}\_k}$ and a penalty per radian for component $k = \textrm{n, t}$ of the orientation error $e_{\textrm{ori\_}k}$.
$r_{\textrm{maintain}\_k}$ is the reward when each component $k$ reaches and remains within the threshold $\delta_k$.
$r_{\textrm{success}}$ is a reward when the task is successful.
$r_{\textrm{drop}}$ is the penalty for when the object is out of reach of the fingers (drop), which is set as $0$ to showcase the advantage of the stable action space.
The specific parameters for the reward function in each task are on the project website (\url{https://hathanglongdoan.github.io/Learning_stable_in-grasp_manipulation/}).

\subsection{Results and Analysis}

\begin{figure*}[!b]
\vspace{-2mm}
\centering
\begin{minipage}[c]{0.02\textwidth}
(a)
\end{minipage}
\begin{minipage}[c]{0.6\textwidth}
\centering
\hspace{4mm}
\includegraphics[width=0.85\linewidth]{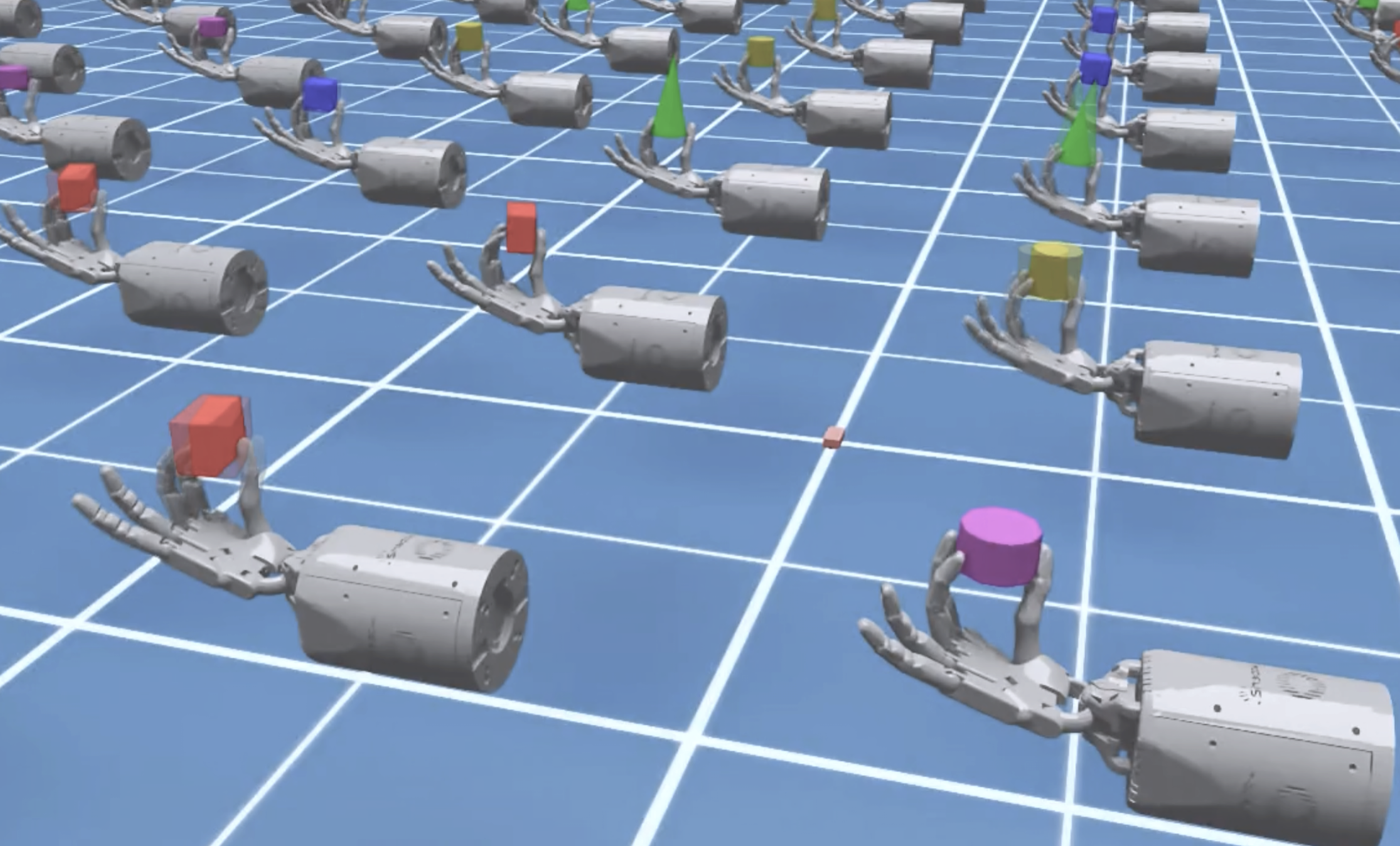}
\end{minipage}
\begin{minipage}[c]{0.30\textwidth}
\renewcommand{\arraystretch}{1.3}
\begin{tabular}{|c | c|}
     \hline
        \makecell{Goal} & \makecell{VF reposition}\\\hline
        \makecell{$I$} & \makecell{th, ff, mf}  \\\hline
         \makecell{Gravity} & \makecell{upward}  \\\hline
\end{tabular}
\end{minipage}
\begin{minipage}[c]{0.49\textwidth}
\centering
\includegraphics[width=1.0\linewidth]{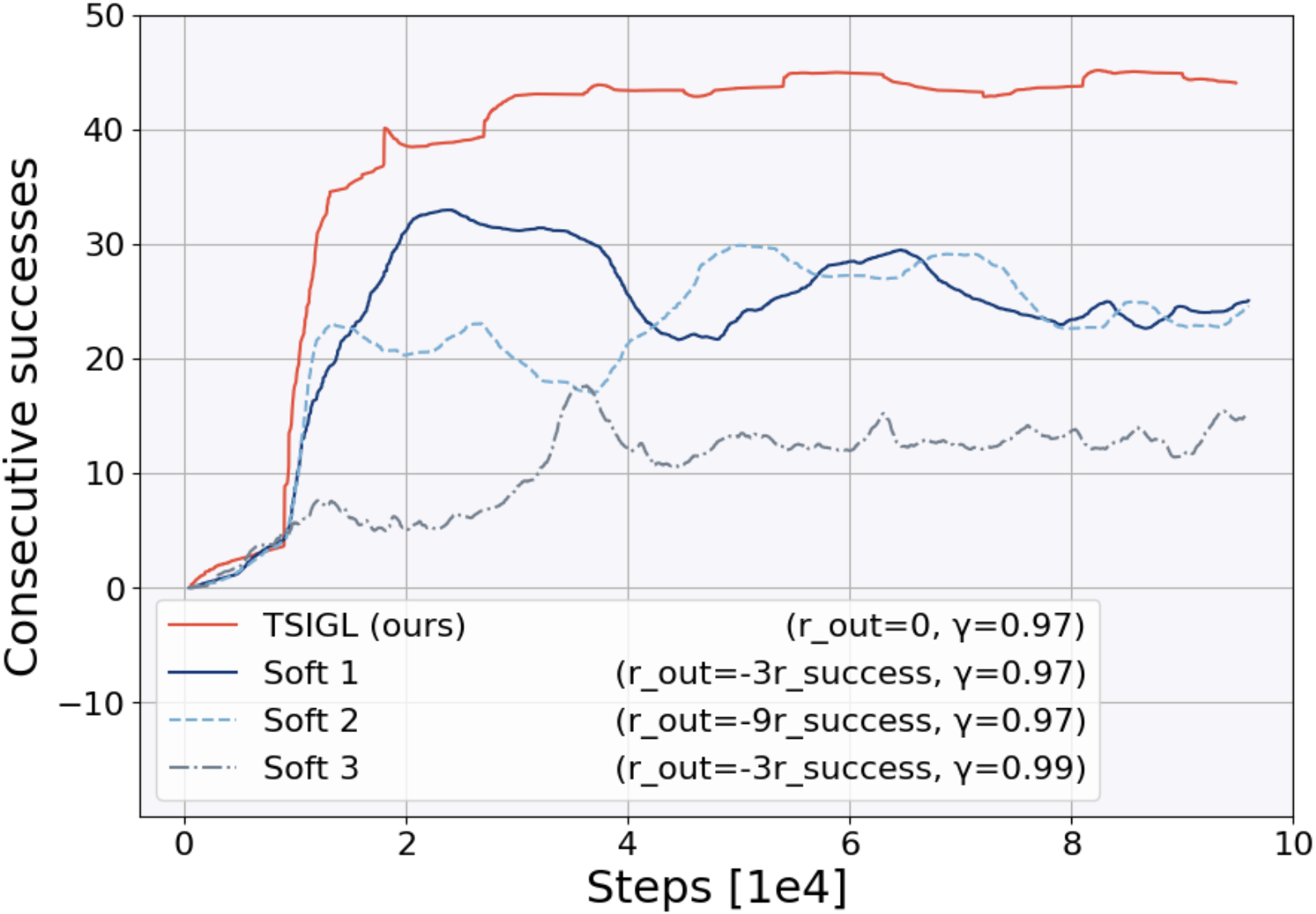}
\end{minipage}
\begin{minipage}[c]{0.49\textwidth}
\centering
\renewcommand{\arraystretch}{1.5}
\begin{tabular}{|c|c|c|}
     \hline
        \makecell{Method} & \makecell{Constraint} & \makecell{Object drop}\\\hline
        \makecell{TSIGL (ours)} &  \makecell{\textcolor{RedOrange}{\textit{CBFs} }} 
        & \color{RedOrange}{\textbf{0}} \\\hline
        
         \makecell{Soft 1} &  \makecell{\textcolor{DarkBlue}{penalty} } & \color{DarkBlue}{738}\\\hline
         
         \makecell{Soft 2} &  \makecell{\textcolor{LightBlue}{penalty} } & \color{LightBlue}{467}\\\hline
         
         \makecell{Soft 3} &  \makecell{\textcolor{ClearBlue}{penalty} } & \color{ClearBlue}{248}\\\hline
\end{tabular}
\end{minipage} \\
\vspace{1mm}\ \ \  \ \ \ \ \ \ \ \ \ \ (b) \ \ \ \ \ \ \ \ \ \ \ \ \ \ \ \ \ \ \ \ \ \ \ \ \ \ \ \ \ \ \ \ \ \ \ \ \ \ \ \ \ \ \ \ \ \ \ \ \ \ \ \ \ \ (c)  \ \ \ \ \ \ \ \ \ \ \
\vspace{-2mm}
\caption{Learning \textit{Skill 2}: (a) Our TSIGL framework learning in-grasp position manipulation skill, (b) sample efficiency comparison with other soft constraint techniques, and (c) number of times an object is dropped during RL.}
\label{learn2}
\end{figure*}

\begin{figure}[!t]
\centering
\begin{minipage}[c]{0.2\textwidth}
\vspace{6mm} (a) \\ \\ \\ \\ \\ \\ \\ \\ \\ \\ (b)
\end{minipage}
\begin{minipage}[c]{0.79\textwidth}
\includegraphics[width=0.85\linewidth]{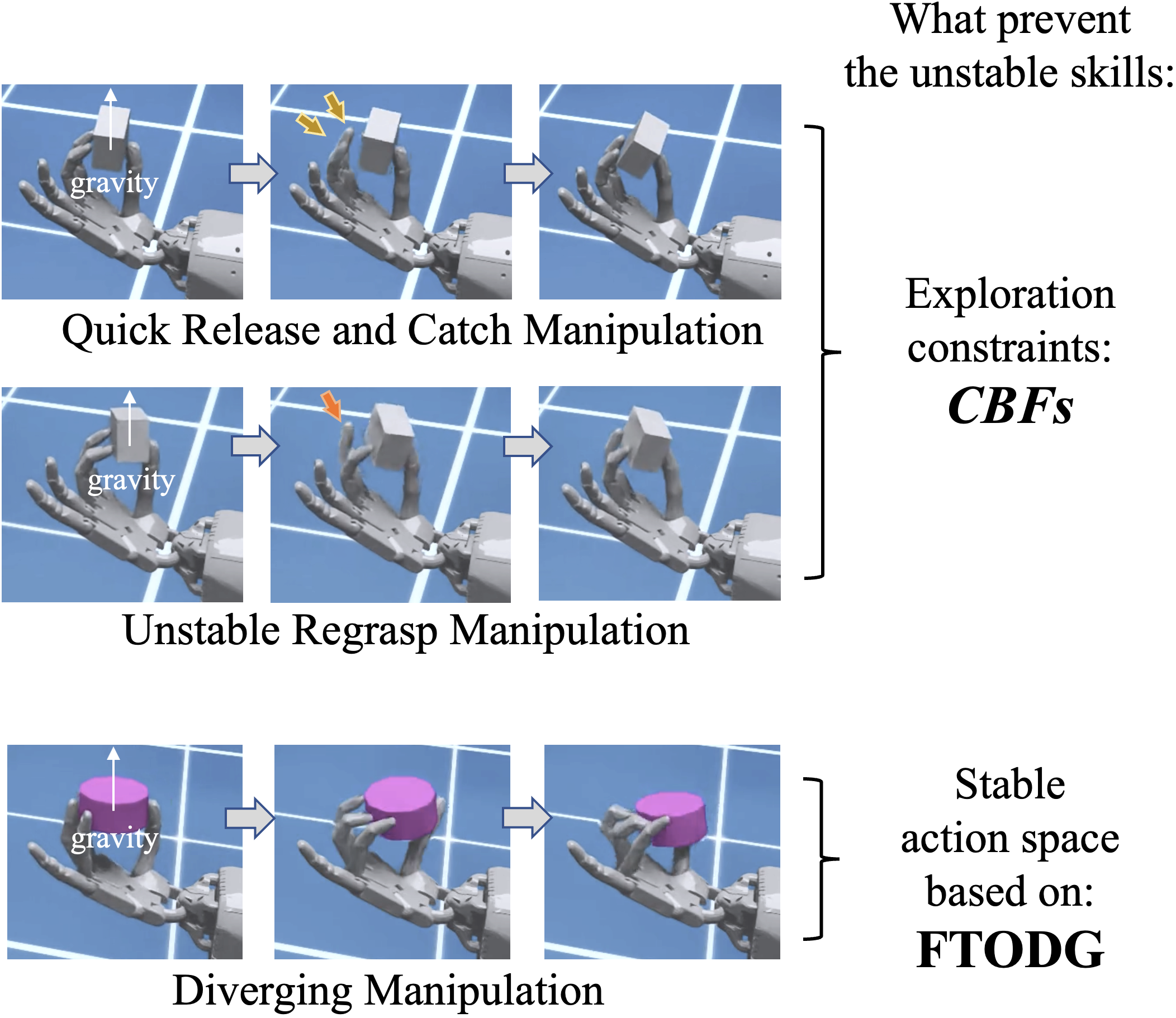}
\end{minipage}
\caption{Unstable skills when (a) softening the implementation requirement constraints, then (b) removing FTODG, confirming the importance of the TSIGL framework design choice.}
\label{deploy2}
\vspace{-3mm}
\end{figure}

\subsubsection{FTODG Stability}
\textit{Force/torque equilibrium grasp} is applied to execute \textit{Skill 1}, and the hand--object system is stabilized.
We then apply external forces to the grasped objects.
The external force range the controller can withstand is obtained by increasing their magnitude until fingertip--object slippage is observed (below 8.6 N for our constant values of $f_{\textrm{d}\_i}$ and $C_i$) (see Figure.~\ref{precision_grasp}(a)).
Within this range, the fingertips move with the objects and keep them stable in the hand.
After the forces are removed, the fingertips slowly move the objects back close to the originally stabilized position (see Figure.~\ref{precision_grasp}(b)).

Therefore, applying the FTODG control design to achieve object stability, as in our TSIGL framework, is effective.

\vspace{1mm}
\subsubsection{CBFs are Necessary to Constrain the Stable Action Space} \label{skill2}
An ablation study of the TSIGL framework is conducted during \textit{Skill 2} learning. We study the effect of softening the implementation requirements for the repositioning force by removing the strict CBF filter and replacing it with a soft penalty.
A success threshold of $\delta = 2.5$mm and a time limit of $T = 4$ seconds are used.
The consecutive successes per environment are recorded at episode resets, and their moving average is plotted (a common evaluation metric for dexterous manipulation RL \cite{openAI}).

\begin{itemize}
\item TSIGL framework: The penalty for dropping the object is set as $r_\textrm{drop} = 0$.
However, because our framework strictly constrains the stable action space using CBFs, the object drop count during exploration is 0 (see Figure.~\ref{learn2}(c)).
The episode resets only when the manipulation goal is not reached in time. Therefore, `consecutive success' is low in the first 8000 steps, then spikes to approximately 39 and converges to 44 at approximately 40000 steps (shown as an orange line in Figure.~\ref{learn2}(b)).

\item Soft 1: Replacing CBFs with penalty results in learning divergence after quickly reaching approximately 33 consecutive successes, or approximately 77\% of our TSIGL framework's result, and does not result in the same training time (shown as a dark blue line in Figure.~\ref{learn2}(b)). Compared with our method, the object also is also dropped more (see Figure.~\ref{learn2}(b)).

\item Soft 2: In an attempt to reduce the object drop count, we increase the penalty $r_{\textrm{drop}}$ by 3 times the original value.
This method reduces the object drop count (see Figure.~\ref{learn2}(c)), but the divergence problem remains (shown as a light blue dashed line in Figure.~\ref{learn2}(b)).

\item Soft 3: We also let the agent value the future reward more by increasing the discount factor $\gamma$, which stabilizes the success metric but slows learning (shown as a gray blue line in Figure.~\ref{learn2}(b)).
Eventually, this approach results in approximately 15 consecutive successes at the end of the training time, or approximately 35\% of our TSIGL framework's result.
\end{itemize}

With respect to the cause of learning divergence, we find two additional skills that are not seen when learning with the TSIGL framework: `quick release and catch manipulation' and `unstable regrasp manipulation' (see Figure.~\ref{deploy2}(a)).
These unstable skills bypass the loophole in the reward function by taking actions that are not stable, as defined in section~\ref{theoretical_stability}, but do not cause obvious slippage or drop right away.
This is one of the causes of the long-tail problem in RL.

Therefore, using \textit{CBFs} to strictly follow the FTODG implementation requirements is necessary to constrain the stable action space.

\begin{figure*}[!t]
\centering
\begin{minipage}[c]{0.02\textwidth}
(a)
\end{minipage}
\begin{minipage}[c]{0.6\textwidth}
\centering
\hspace{4mm}
\includegraphics[width=0.85\linewidth]{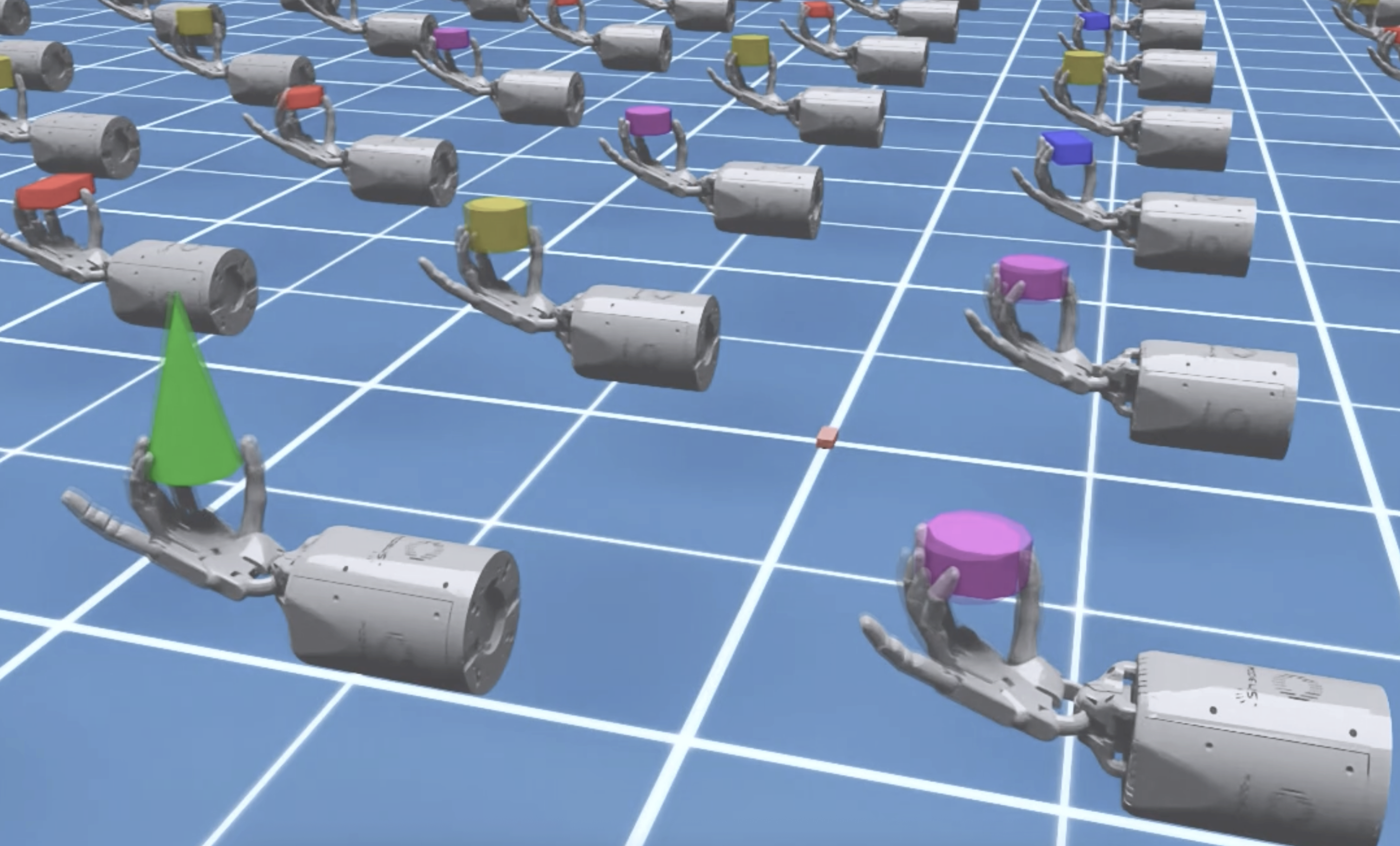}
\end{minipage}
\vspace{2mm}
\begin{minipage}[c]{0.30\textwidth}
\renewcommand{\arraystretch}{1.6}
\begin{tabular}{|c | c|}
     \hline
        \makecell{Goal} & \makecell{Object\\reorientation}\\\hline
        \makecell{$I$} & \makecell{th, ff, mf, rf}  \\\hline
         \makecell{Gravity} & \makecell{upward}  \\\hline
\end{tabular}
\end{minipage}
\hspace{0.3mm}
\begin{minipage}[c]{0.49\textwidth}
\centering
\includegraphics[width=0.99\linewidth]{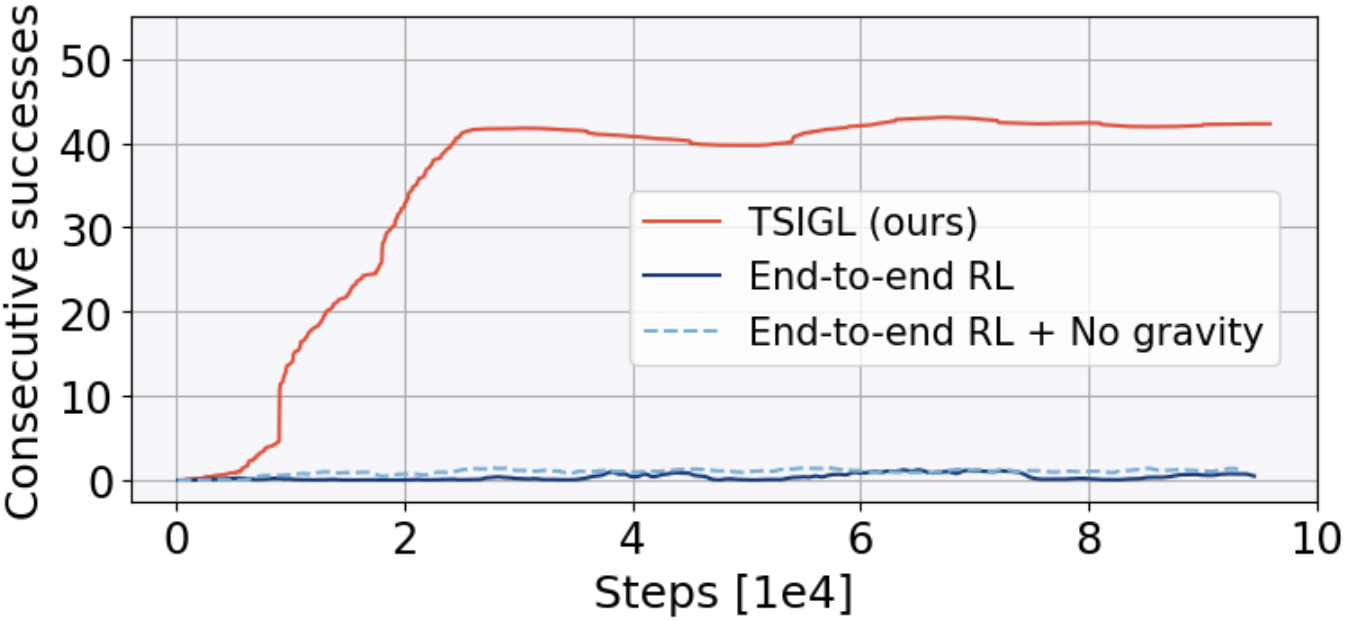}
\centering
\end{minipage}
\begin{minipage}[c]{0.49\textwidth}
\renewcommand{\arraystretch}{1.1}
\begin{tabular}{|c|c|c|}
     \hline
     \makecell{Learning methods} & \makecell{Object drop} \\\hline
        \makecell{TSIGL (ours)\ \ \ \ \ \ \ \ \ \ \ \ \ \ } & \color{RedOrange}{\textbf{0}}\\
         \makecell{End-to-end \cite{IsaacLab, openAI} \ \ \ \ \ } & \color{DarkBlue}{3138}\\
         \makecell{End-to-end + No gravity} &  \color{LightBlue}{1282}\\\hline
\end{tabular}
\end{minipage}\hfill \newline
\\
(b) $\delta = 0.1$ rad, $T = 4$ seconds 
\\
\vspace{2mm}
\begin{minipage}[c]{0.49\textwidth}
\centering
\includegraphics[width=0.99\linewidth]{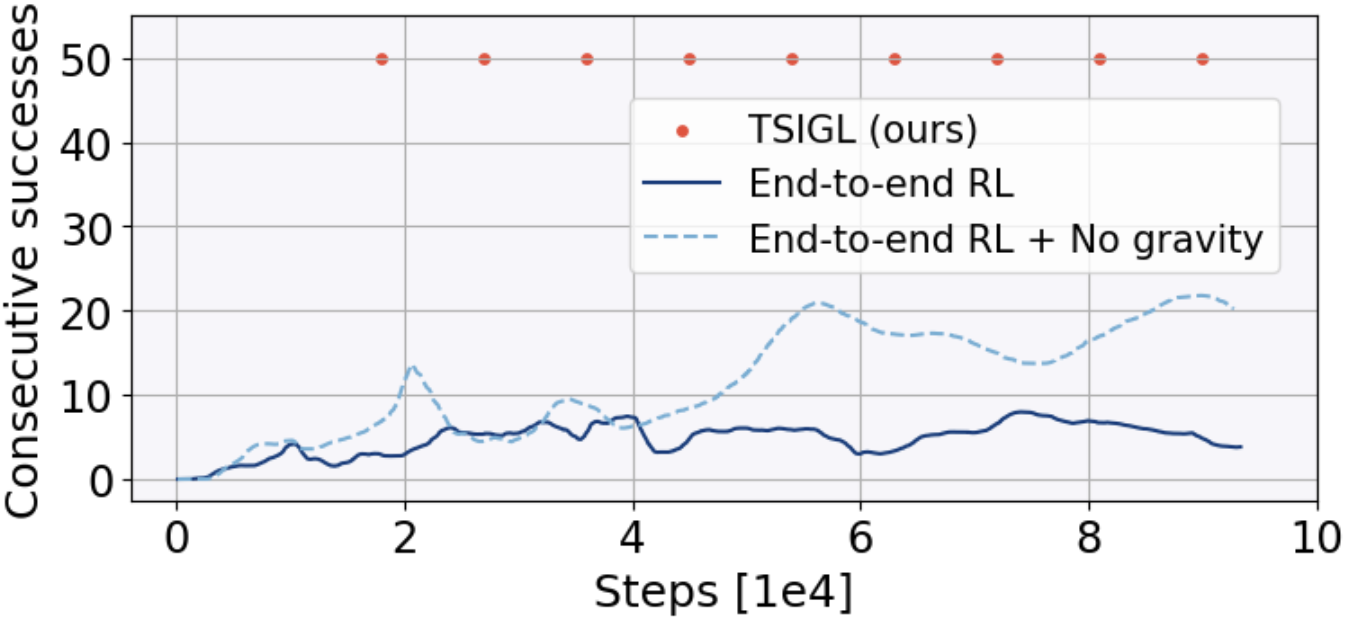}
\end{minipage}
\begin{minipage}[c]{0.49\textwidth}
\vspace{1mm}
\renewcommand{\arraystretch}{1.1}
\hspace{1mm}
\begin{tabular}{|c|c|c|}
     \hline
     \makecell{Learning methods} & \makecell{Object drop} \\\hline
        \makecell{TSIGL (ours)\ \ \ \ \ \ \ \ \ \ \ \ \ \ } & \color{RedOrange}{\textbf{0}}\\
         \makecell{End-to-end \cite{IsaacLab, openAI} \ \ \ \ \ } & \color{DarkBlue}{12317}\\
         \makecell{End-to-end + No gravity} &  \color{LightBlue}{4880}\\\hline
\end{tabular}
\end{minipage}\hfill \newline
\\
(c) $\delta = 0.4$ rad, $T = 5$ seconds
\vspace{-1mm}
\caption{Learning \textit{Skill 3}: (a) Our TSIGL framework for learning in-grasp orientation manipulation skill; (b) main task, and (c) simplified task's RL performance compared to other learning methods.}
\label{learn3}
\vspace{-3mm}
\end{figure*}

\subsubsection{Exploration Within the Stable Action Space is Sample Efficient} \label{skill3}
In \textit{Skill 3} learning, an ablation study of removing the FTODG's guidance to construct the stable action space is conducted.
We compare TSIGL with the baseline end-to-end learning method available in NVIDIA Isaac Lab \cite{orbit, IsaacLab} based on OpenAI \textit{et al.}'s work \cite{openAI}.
The curriculum to gradually introduce gravity \cite{DLR1} is also applied to one end-to-end learning environment.
Because our training time is short, zero gravity is applied throughout.
A success threshold of $\delta = 0.1$ rad and a time limit of $T = 4$ seconds are used.

\begin{itemize}
\item TSIGL framework:
Owing to the constrained stable action space, the object drop count is 0 during exploration (see Figure.~\ref{learn3}(c)), and more than 42 consecutive successes are reached after 25000 steps (shown as an orange line in Figure.~\ref{learn3}(b)).

\item End-to-end learning: The environment with gravity reaches 0.8 consecutive manipulation successes (shown as a dark blue line in Figure.~\ref{learn3}(b)).
Zero gravity promotes learning by reducing more than half of the object drop cases (see Figure.~\ref{learn3}(c)) and reaches approximately 2.1 consecutive successes on average (shown as a light blue dashed line in Figure.~\ref{learn3}(b)).
During this stage, `quick release and catch' and `unstable regrasp' skills are explored in both methods.
They also execute `diverging manipulation' (see Figure.~\ref{deploy2}(b)) and frequently drop objects.

\item Simplified task: We also conduct a simplified task with increased success tolerance ($\delta = 0.4$ rad) and time limit ($T = 5$ seconds):
Learning from scratch with gravity reaches approximately 8 consecutive successes after 9000 steps (shown as a dark blue line in Figure.~\ref{learn3}(c)).
The effect of the gravity curriculum is clearly shown when it reduces the object drop cases by 2.5 times and reaches approximately 22 consecutive successes (shown as a light blue dashed line in Figure.~\ref{learn3}(c)).
However, with such a high success tolerance, the TSIGL framework does not fail, and the episodes only reset after a maximum consecutive number of successes of 50 (highlighted using orange markers in Figure.~\ref{learn3}(c)).

\end{itemize}

Therefore, compared with baseline end-to-end learning methods, constrained exploration within the stable action space in our TSIGL framework is more sample efficient.

\begin{figure}[!b]
\vspace{-3mm}
\centering
 \ \ \ \ \ \ \ \ \ \ \ \ \ \ \ \ \ \ \ \ \ \ (a)\ \ \ \ \ \ \ \ \ \ (b)\ \ \ \ \ \ \ \ \ \ (c)\ \ \ \ \ \ \ \ \ \ (d) \\\vspace{1mm}
\renewcommand{\arraystretch}{1.3}
\begin{tabular}{|c|c|c|c|c|}
\hline\makecell{\textit{Skill 2}}&\makecell{\includegraphics[width=0.1\linewidth]{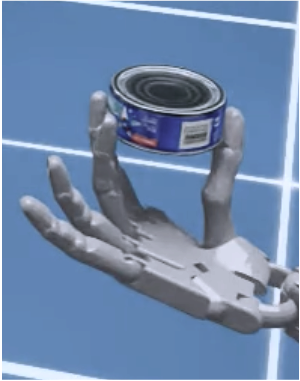}}&\makecell{\includegraphics[width=0.1\linewidth]{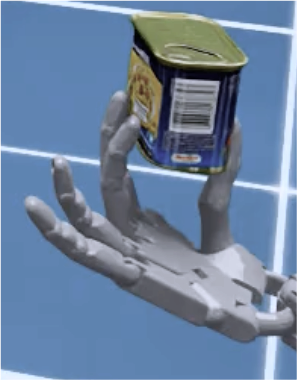}}&\makecell{\includegraphics[width=0.1\linewidth]{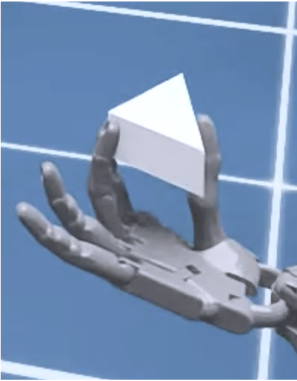}}&\makecell{\includegraphics[width=0.1\linewidth]{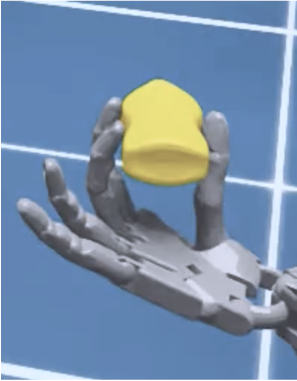}}\\\hline\hline
        \makecell{FTODG} &20.3&18.0&18.1&14.3\\
        \makecell{TSIGL (ours)} &\textbf{48.3}&\textbf{41.5}&\textbf{43.4}&\textbf{36.5}\\\hline \hline
\end{tabular}
\centering
\begin{tabular}{|c|c|c|c|c|}
        \makecell{\textit{Skill 3}}&\makecell{\includegraphics[width=0.1\linewidth]{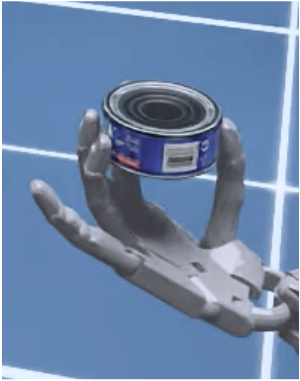}}&\makecell{\includegraphics[width=0.1\linewidth]{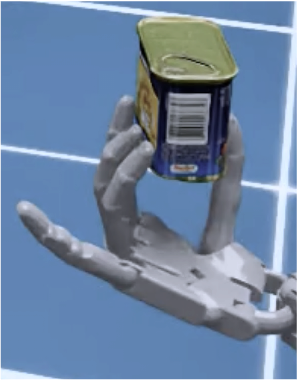}}&\makecell{\includegraphics[width=0.1\linewidth]{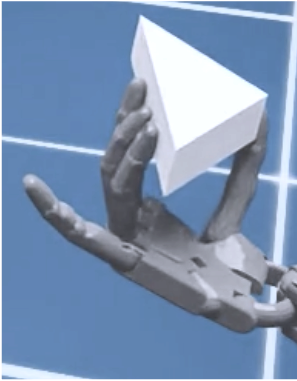}}&\makecell{\includegraphics[width=0.1\linewidth]{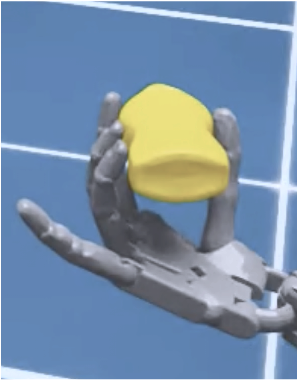}}\\\hline\hline
        \makecell{FTODG} &23.6&25.2&18.5&33.3\\
        \makecell{TSIGL (ours)} &\textbf{42.6}&\textbf{39.2}&\textbf{33.4}&\textbf{46.4}\\\hline
\end{tabular}
\caption{Average consecutive successes when FTODG's controllers and TSIGL's learned policy are deployed on a (a) tuna can, (b) spam can, (c) triangular prism, and (d) mustard bottle.}
\label{deploy3}
\end{figure}

\begin{figure}[!b]
\vspace{-1mm}
\centering
\includegraphics[width=0.665\linewidth]{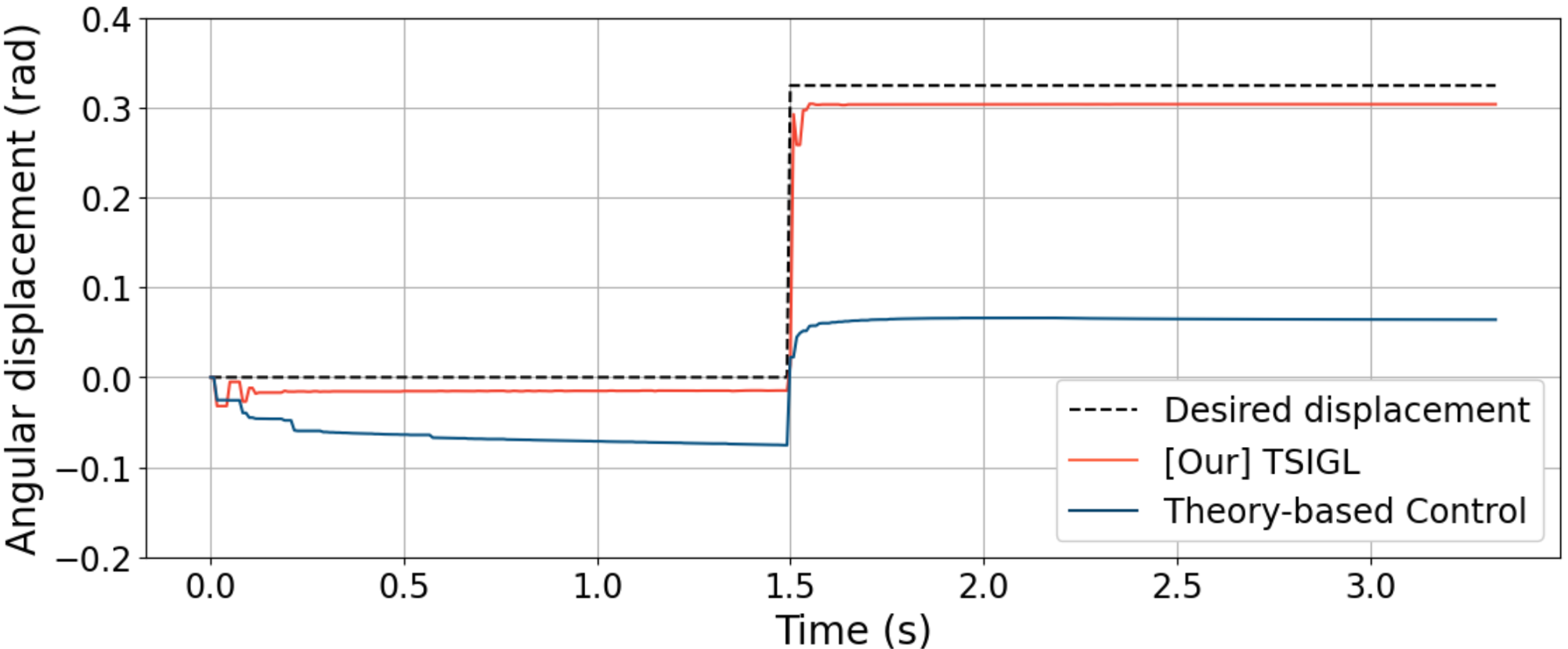}
\vspace{-2mm}
\caption{Orientation manipulation step response data showing that the TSIGL framework outperforms FTODG in accuracy.}
\label{step_response}
\end{figure}

\subsubsection{Improved Accuracy with RL} 

The learned policies of the TSIGL framework for \textit{Skill 2} and \textit{Skill 3} are deployed and compared to those of the controllers with fixed parameters studied in FTODG.

We record the number of consecutive successes the first time each of the 96 environments resets. The average value for each object highlights the accuracy improvement of using the TSIGL framework compared with applying the theory directly:
The values of \textit{Skill 2} increase from 130 to 155\% (see Figure.~\ref{deploy2}(a)), and the values of \textit{Skill 3} increase from 39 to 81\% (see Figure.~\ref{deploy2}(b)).
Step response data of one orientation manipulation task on a tuna can be plotted (see Figure.~\ref{step_response}), showcasing the improvement in manipulation accuracy owing to the reorientation moment being explored with guided RL.

Therefore, using RL actions to address the accuracy limitations of the control design of FTODG, as in our TSIGL framework, is effective.

\section{Conclusion}
\subsection{Summary}
This study proposes an effective method to guide learning stable in-grasp manipulation with prior scientific knowledge studied in FTODG \cite{hcr_theory}. By utilizing its strength of stability, its accuracy limitations can be optimized while strictly following its implementation requirements using \textit{control barrier functions}.
Object dropping is prevented, and compared with baseline end-to-end RL methods, the sample efficiency of skill learning is significantly improved.

\subsection{Limitations}
First, our work is performed on a well-recognized simulation in the field \cite{orbit, IsaacLab} but not in the real world.
However, we showcase the high potential of our TSIGL framework, with significant improvements in various evaluation metrics compared with baseline methods.
Second, our approach constrains the RL action space to obtain stability for a predefined skill, which limits the ability to explore.
However, a trade-off can be made by softening the constraints, as shown in this paper.
Finally, other important aspects of manipulation, such as grasp quality and hand--object workspace analysis, are not discussed.
In the future, we will extend our work to include these contents and realize other dexterous skills in the real world.

\section*{Disclosure statement}
No potential conflict of interest was reported by the author(s).

\section*{Funding}
This work was supported by JSPS KAKENHI (Grant Numbers JP25KJ1913 and JP24H00726).

\section*{Notes on contributors}
\textbf{Ha Thang Long Doan} received the B.Eng. degree in mechanical and aerospace engineering, and M.Eng. in mechanical engineering from Kyushu University, Fukuoka, Japan, in 2022 and 2024, respectively. He is currently pursuing a Ph.D. degree in mechanical engineering at Kyushu University. His research interests include dexterous manipulation, data-driven control, and reinforcement learning.\\ \\
\textbf{Hikaru Arita} received the B.Eng., M.Eng., and Ph.D. degrees in engineering from The University of Electro-Communications, Tokyo, Japan, in 2012, 2014, and 2019, respectively. He worked for OMRON Corporation, Kyoto, Japan, from 2014 to 2016, and Ritsumeikan University, where he was an Assistant Professor, from 2019 to 2022. In 2022, he joined Kyushu University, Fukuoka, Japan, as an Assistant Professor. He has been an Associate Professor with Kyushu University since 2024. His research interests include proximity sensors, sensor-based control, soft robotics, force control, bipedal robots, and manipulation.\\ \\ 
\textbf{Kazuto Nakashima} received the M.Eng. and Ph.D. degrees from Kyushu University, Japan, in 2017 and 2020, respectively. He is currently an Associate Professor in the Department of Interdisciplinary Informatics at Kyushu University, with a concurrent appointment in the Department of Mechanical Engineering. Previously, he was a Postdoctoral Researcher from 2021 to 2023 and an Assistant Professor from 2023 to 2024 in the Department of Advanced Information Technology at Kyushu University. His research interests include computer vision and machine learning for robotics, with a particular focus on scene understanding, generative modeling, and state estimation.\\ \\
\textbf{Kenji Tahara} received the B.S. degree in mechanical engineering, the M.S. degree in information science and systems, and the Ph.D. degree in robotics from Ritsumeikan University, Japan, in 1998, 2000, and 2003, respectively. From 2003 to 2007, he was a Research Scientist with the Bio-Mimetic Control Research Center, RIKEN. In 2007, he joined Kyushu University as a tenure-tracked Associate Professor. In 2011, he was an Associate Professor in the Department of Mechanical Engineering at Kyushu University, where he has been a Full Professor since 2020. His current research interests include multi-fingered hands, manipulation, force control, bipedal robots, and soft robotics.


\bibliographystyle{IEEEtran}
\bibliography{TIRL.bib}

@inproceedings{hcr_object_ori_feedback,
	author = {Kawamura, Akihiro and Tahara, Kenji and Kurazume, Ryo and Hasegawa, Tsutomu},
	booktitle = {Proceedings of the IEEE/RSJ International Conference on Intelligent Robots and Systems (IROS)},
	doi = {10.1109/IROS.2012.6385589},
	pages = {4797-4803},
	title = {Robust visual servoing for object manipulation with large time-delays of visual information},
	year = {2012}}

@article{hcr_long,
	author = {Doan, Ha Thang Long and Arita, Hikaru and Tahara, Kenji},
	da = {2025/11/19},
	doi = {10.1186/s40648-025-00329-y},
	id = {Doan2025},
	isbn = {2197-4225},
	journal = {ROBOMECH Journal},
	number = {1},
	pages = {39},
	title = {Enabling external sensorless in-hand object position manipulation by linkage-based underactuated hands with mechanical stoppers},
	ty = {JOUR},
	volume = {12},
	year = {2025}}

@article{hcr_tactile_manipulation,
	author = {Choi, Seung-hyun and Tahara, Kenji},
	da = {2020/03/12},
	doi = {10.1186/s40648-020-00162-5},
	id = {Choi2020},
	isbn = {2197-4225},
	journal = {ROBOMECH Journal},
	number = {1},
	pages = {14},
	title = {Dexterous object manipulation by a multi-fingered robotic hand with visual-tactile fingertip sensors},
	ty = {JOUR},
	volume = {7},
	year = {2020}}

@inproceedings{RLphysics_control,
	author = {Margolis, Gabriel B and Chen, Tao and Paigwar, Kartik and Fu, Xiang and Kim, Donghyun and Kim, Sang bae and Agrawal, Pulkit},
	booktitle = {Proceedings of the 5th Conference on Robot Learning},
	editor = {Faust, Aleksandra and Hsu, David and Neumann, Gerhard},
	month = {08--11 Nov},
	pages = {1025--1034},
	publisher = {PMLR},
	series = {Proceedings of Machine Learning Research},
	title = {Learning to Jump from Pixels},
	volume = {164},
	year = {2022}}

@inproceedings{RLphysics_reward,
	author = {Zipeng Fu and Ashish Kumar and Jitendra Malik and Deepak Pathak},
	booktitle = {Conference on Robot Learning},
	title = {Minimizing Energy Consumption Leads to the Emergence of Gaits in Legged Robots},
	year = {2021}}

@article{safeRL,
	author = {Brunke, Lukas and Greeff, Melissa and Hall, Adam W. and Yuan, Zhaocong and Zhou, Siqi and Panerati, Jacopo and Schoellig, Angela P.},
	doi = {https://doi.org/10.1146/annurev-control-042920-020211},
	issn = {2573-5144},
	journal = {Annual Review of Control, Robotics, and Autonomous Systems},
	number = {Volume 5},
	pages = {411-444},
	publisher = {Annual Reviews},
	title = {Safe Learning in Robotics: From Learning-Based Control to Safe Reinforcement Learning},
	type = {Journal Article},
	volume = {5},
	year = {2022}}

@article{physicsRL,
	author = {E{\ss}er, Julian and Bach, Nicolas and Jestel, Christian and Urbann, Oliver and Kerner, S{\"o}ren},
	doi = {10.1109/MRA.2022.3207664},
	journal = {IEEE Robotics and Automation Magazine},
	number = {2},
	pages = {67-85},
	title = {Guided Reinforcement Learning: A Review and Evaluation for Efficient and Effective Real-World Robotics},
	volume = {30},
	year = {2023}}

@article{stable_learning,
	author = {Tao, Lingfeng and Zhang, Jiucai and Zhang, Xiaoli},
	doi = {10.1109/LRA.2025.3530323},
	journal = {IEEE Robotics and Automation Letters},
	number = {3},
	pages = {2407-2413},
	title = {Stable In-Hand Manipulation With Finger-Specific Multi-Agent Shadow Critic Consensus and Information Sharing},
	volume = {10},
	year = {2025}}

@book{hcr_theory,
	author = {Arimoto, S.},
	isbn = {9781848000636},
	lccn = {2007941071},
	publisher = {Springer London},
	title = {Control Theory of Multi-fingered Hands: A Modelling and Analytical--Mechanics Approach for Dexterity and Intelligence},
	year = {2008}}

@book{robot_book,
	author = {Lynch, {Kevin M} and Park, {Frank C.}},
	isbn = {978-1107156302},
	language = {English (US)},
	publisher = {Cambridge Univeristy Press},
	title = {Modern Robotics: Mechanics, Planning, and Control},
	year = {2017}}

@inproceedings{power_grasp,
	author = {Bicchi, A.},
	booktitle = {Proceedings of the IEEE/RSJ International Workshop on Intelligent Robots and Systems},
	doi = {10.1109/IROS.1991.174559},
	pages = {691-697 vol.2},
	title = {Analysis and control of power grasping},
	year = {1991}}

@article{slide_regrasp,
	author = {Cole, A.A. and Hsu, P. and Sastry, S.S.},
	doi = {10.1109/70.127238},
	journal = {IEEE Transactions on Robotics and Automation},
	number = {1},
	pages = {42-52},
	title = {Dynamic control of sliding by robot hands for regrasping},
	volume = {8},
	year = {1992}}

@article{position_control,
	author = {Zribi, Mohamed and Chen, Jun and Mahmoud, Magdi S.},
	da = {1999/02/01},
	doi = {10.1023/A:1008097528538},
	id = {Zribi1999},
	isbn = {1573-0409},
	journal = {Journal of Intelligent and Robotic Systems},
	number = {2},
	pages = {125--149},
	title = {Coordination and Control of Multi-fingered Robot Hands with Rolling and Sliding Contacts},
	ty = {JOUR},
	volume = {24},
	year = {1999}}

@inproceedings{DLR2,
	author = {Pitz, Johannes and R{\"o}stel, Lennart and Sievers, Leon and Burschka, Darius and B{\"a}uml, Berthold},
	booktitle = {Proceedings of the IEEE/RSJ International Conference on Intelligent Robots and Systems (IROS)},
	doi = {10.1109/IROS58592.2024.10802864},
	pages = {13112-13119},
	title = {Learning a Shape-Conditioned Agent for Purely Tactile In-Hand Manipulation of Various Objects},
	year = {2024}}

@inproceedings{residual,
	author = {Johannink, Tobias and Bahl, Shikhar and Nair, Ashvin and Luo, Jianlan and Kumar, Avinash and Loskyll, Matthias and Ojea, Juan Aparicio and Solowjow, Eugen and Levine, Sergey},
	booktitle = {Proceedings of the International Conference on Robotics and Automation (ICRA)},
	doi = {10.1109/ICRA.2019.8794127},
	pages = {6023-6029},
	title = {Residual Reinforcement Learning for Robot Control},
	year = {2019}}

@misc{GAE,
	archiveprefix = {arXiv},
	author = {John Schulman and Philipp Moritz and Sergey Levine and Michael Jordan and Pieter Abbeel},
	eprint = {1506.02438},
	primaryclass = {cs.LG},
	title = {High-Dimensional Continuous Control Using Generalized Advantage Estimation},
	url = {https://arxiv.org/abs/1506.02438},
	year = {2018}}

@misc{PPO,
	archiveprefix = {arXiv},
	author = {John Schulman and Filip Wolski and Prafulla Dhariwal and Alec Radford and Oleg Klimov},
	eprint = {1707.06347},
	primaryclass = {cs.LG},
	title = {Proximal Policy Optimization Algorithms},
	url = {https://arxiv.org/abs/1707.06347},
	year = {2017}}

@inproceedings{CBFs,
	author = {Ames, Aaron D. and Coogan, Samuel and Egerstedt, Magnus and Notomista, Gennaro and Sreenath, Koushil and Tabuada, Paulo},
	booktitle = {2019 18th European Control Conference (ECC)},
	doi = {10.23919/ECC.2019.8796030},
	pages = {3420-3431},
	title = {Control Barrier Functions: Theory and Applications},
	year = {2019}}

@article{RL_CBFs,
	abstractnote = {&lt;p&gt;Reinforcement Learning (RL) algorithms have found limited success beyond simulated applications, and one main reason is the absence of safety guarantees &lt;em&gt;during&lt;/em&gt; the learning process. Real world systems would realistically fail or break before an optimal controller can be learned. To address this issue, we propose a controller architecture that combines (1) a model-free RL-based controller with (2) model-based controllers utilizing control barrier functions (CBFs) and (3) online learning of the unknown system dynamics, in order to ensure safety during learning. Our general framework leverages the success of RL algorithms to learn high-performance controllers, while the CBF-based controllers both &lt;em&gt;guarantee&lt;/em&gt; safety and &lt;em&gt;guide&lt;/em&gt; the learning process by constraining the set of explorable polices. We utilize Gaussian Processes (GPs) to model the system dynamics and its uncertainties.&lt;/p&gt; &lt;p&gt;Our novel controller synthesis algorithm, RL-CBF, guarantees safety with high probability during the learning process, regardless of the RL algorithm used, and demonstrates greater policy exploration efficiency. We test our algorithm on (1) control of an inverted pendulum and (2) autonomous carfollowing with wireless vehicle-to-vehicle communication, and show that our algorithm attains much greater sample efficiency in learning than other state-of-the-art algorithms &lt;em&gt;and&lt;/em&gt; maintains safety during the entire learning process.&lt;/p&gt;},
	author = {Cheng, Richard and Orosz, G{\'a}bor and Murray, Richard M. and Burdick, Joel W.},
	doi = {10.1609/aaai.v33i01.33013387},
	journal = {Proceedings of the AAAI Conference on Artificial Intelligence},
	month = {Jul.},
	number = {01},
	pages = {3387-3395},
	title = {End-to-End Safe Reinforcement Learning through Barrier Functions for Safety-Critical Continuous Control Tasks},
	volume = {33},
	year = {2019}}

@webpage{IsaacLab,
	url = {https://isaac-sim.github.io/IsaacLab/v2.3.0}}

@article{Orbit,
	author = {Mittal, Mayank and Yu, Calvin and Yu, Qinxi and Liu, Jingzhou and Rudin, Nikita and Hoeller, David and Yuan, Jia Lin and Singh, Ritvik and Guo, Yunrong and Mazhar, Hammad and Mandlekar, Ajay and Babich, Buck and State, Gavriel and Hutter, Marco and Garg, Animesh},
	doi = {10.1109/LRA.2023.3270034},
	journal = {IEEE Robotics and Automation Letters},
	number = {6},
	pages = {3740-3747},
	title = {Orbit: A Unified Simulation Framework for Interactive Robot Learning Environments},
	volume = {8},
	year = {2023}}

@article{superposition,
	author = {Arimoto, S. and Tahara, K. and Yamaguchi, M. and Nguyen, P.T.A. and Han, M.-Y.},
	doi = {10.1017/S0263574700002939},
	journal = {Robotica},
	number = {1},
	pages = {21--28},
	publisher = {Cambridge University Press},
	title = {Principles of superposition for controlling pinch motions by means of robot fingers with soft tips},
	volume = {19},
	year = {2001}}

@misc{openAI_rubik,
	archiveprefix = {arXiv},
	author = {OpenAI and Ilge Akkaya and Marcin Andrychowicz and Maciek Chociej and Mateusz Litwin and Bob McGrew and Arthur Petron and Alex Paino and Matthias Plappert and Glenn Powell and Raphael Ribas and Jonas Schneider and Nikolas Tezak and Jerry Tworek and Peter Welinder and Lilian Weng and Qiming Yuan and Wojciech Zaremba and Lei Zhang},
	eprint = {1910.07113},
	primaryclass = {cs.LG},
	title = {Solving Rubik's Cube with a Robot Hand},
	url = {https://arxiv.org/abs/1910.07113},
	year = {2019}}

@inproceedings{DLR1,
	author = {Sievers, Leon and Pitz, Johannes and B{\"a}uml, Berthold},
	booktitle = {Proceedings of the International Conference on Robotics and Automation (ICRA)},
	doi = {10.1109/ICRA46639.2022.9812093},
	pages = {2745-2751},
	title = {Learning Purely Tactile In-Hand Manipulation with a Torque-Controlled Hand},
	year = {2022}}

@article{review_analytical-based,
	author = {Ryuta Ozawa and Kenji Tahara},
	doi = {10.1080/01691864.2017.1365011},
	journal = {Advanced Robotics},
	number = {19-20},
	pages = {1030-1050},
	publisher = {Taylor & Francis},
	title = {Grasp and dexterous manipulation of multi-fingered robotic hands: a review from a control view point},
	volume = {31},
	year = {2017}}

@article{tactile_learning_manipulation,
	author = {Funabashi, Satoshi and Isobe, Tomoki and Hongyi, Fei and Hiramoto, Atsumu and Schmitz, Alexander and Sugano, Shigeki and Ogata, Tetsuya},
	doi = {10.1109/LRA.2022.3142417},
	journal = {IEEE Robotics and Automation Letters},
	number = {2},
	pages = {2102-2109},
	title = {Multi-Fingered In-Hand Manipulation With Various Object Properties Using Graph Convolutional Networks and Distributed Tactile Sensors},
	volume = {7},
	year = {2022}}

@article{openAI,
	author = {OpenAI and Marcin Andrychowicz and Bowen Baker and Maciek Chociej and Rafal J{\'o}zefowicz and Bob McGrew and Jakub Pachocki and Arthur Petron and Matthias Plappert and Glenn Powell and Alex Ray and Jonas Schneider and Szymon Sidor and Josh Tobin and Peter Welinder and Lilian Weng and Wojciech Zaremba},
	doi = {10.1177/0278364919887447},
	journal = {The International Journal of Robotics Research},
	number = {1},
	pages = {3-20},
	title = {Learning dexterous in-hand manipulation},
	volume = {39},
	year = {2020}}

@inproceedings{hcr_stable_grasping,
	author = {Tahara, Kenji and Arimoto, Suguru and Yoshida, Morio},
	booktitle = {Proceedings of the IEEE/RSJ International Conference on Intelligent Robots and Systems (IROS)},
	doi = {10.1109/IROS.2009.5354563},
	pages = {2257-2263},
	title = {Dynamic force/torque equilibrium for stable grasping by a triple robotic fingers system},
	year = {2009}}

@inproceedings{hcr_virtual_frame,
	author = {Tahara, Kenji and Arimoto, Suguru and Yoshida, Morio},
	booktitle = {Proceedings of the IEEE International Conference on Robotics and Automation},
	doi = {10.1109/ROBOT.2010.5509372},
	pages = {4322-4327},
	title = {Dynamic object manipulation using a virtual frame by a triple soft-fingered robotic hand},
	year = {2010}}

@inproceedings{hcr_regrasp,
	author = {Tahara, Kenji and Maruta, Keigo and Kawamura, Akihiro and Yamamoto, Motoji},
	booktitle = {Proceedings of the IEEE International Conference on Robotics and Automation},
	doi = {10.1109/ICRA.2012.6224681},
	pages = {3252-3257},
	title = {Externally sensorless dynamic regrasping and manipulation by a triple-fingered robotic hand with torsional fingertip joints},
	year = {2012}}

@article{analytical_reorientation,
	author = {Li, Qiang and Meier, Martin and Haschke, Robert and Ritter, Helge and Bolder, Bram  },
	doi = {10.1504/IJMA.2013.052624},
	journal = {International Journal of Mechatronics and Automation (IJMA)},
	month = {01},
	title = {Rotary Object Dexterous Manipulation in Hand: A Feedback-based Method},
	volume = {3},
	year = {2013}}

@webpage{shadow_hand,
	url = {https://shadowrobot.com/dexterous-hand-series}}

@article{review_dexterous_manipulation_PROBLEM,
	author = {Jinda Cui and Jeff Trinkle},
	doi = {10.1126/scirobotics.abd9461},
	journal = {Science Robotics},
	number = {54},
	pages = {eabd9461},
	title = {Toward next-generation learned robot manipulation},
	volume = {6},
	year = {2021}}

@article{review_dexterous_manipulation,
	author = {Aude Billard and Danica Kragic},
	doi = {10.1126/science.aat8414},
	journal = {Science},
	number = {6446},
	pages = {eaat8414},
	title = {Trends and challenges in robot manipulation},
	volume = {364},
	year = {2019}}

\end{document}